\documentclass{article}

\usepackage{microtype}
\usepackage{graphicx}
\usepackage{subcaption}
\usepackage{booktabs} % for professional tables
\usepackage{fontawesome5}

\usepackage{hyperref}

\usepackage[preprint]{icml2026}

\usepackage{amsmath}
\usepackage{amssymb}
\usepackage{mathtools}
\usepackage{amsthm}

\usepackage[capitalize,noabbrev]{cleveref}

\usepackage{hyperref}
\usepackage{url}
\usepackage{color}
\usepackage{colortbl}
\usepackage{caption}
\usepackage{array}
\usepackage{multirow}
\usepackage{booktabs}
\usepackage{pifont}
\usepackage{amssymb}
\usepackage{amsmath}
\usepackage{bm}
\usepackage{tikz}
\definecolor{Gray}{gray}{0.93}
\usepackage{listings}
\usepackage{algorithm}
\usepackage{algorithmic}
\usepackage{enumitem}
\usepackage{tcolorbox}
\tcbuselibrary{listings}
\renewcommand{\algorithmicrequire}{\textbf{Input:}} %Use Input in the format of Algorithm
\renewcommand{\algorithmicensure}{\textbf{Output:}} %UseOutput in the format of Algorithm
\newlength\savewidth\newcommand\shline{\noalign{\global\savewidth\arrayrulewidth
  \global\arrayrulewidth 1pt}\hline\noalign{\global\arrayrulewidth\savewidth}}
\definecolor{deemph}{gray}{0.62}
\definecolor{mgreen}{rgb}{0.19,0.80,0.19}
\newcommand{\gc}[1]{\textcolor{deemph}{#1}}
\newcommand{\videover}{{\textbf{\textsc{VideoVeritas}}}}
\definecolor{selfblue}{RGB}{65,105,225} %{89,138,234}
\definecolor{lightblue}{RGB}{220,230,255}
\theoremstyle{plain}

\theoremstyle{definition}

\theoremstyle{remark}

\usepackage[textsize=tiny]{todonotes}

\icmltitlerunning{\textsc{VideoVeritas}: AI-Generated Video Detection via Perception Pretext Reinforcement Learning}

\begin{document}

\twocolumn[
  \icmltitle{\textsc{VideoVeritas}: AI-Generated Video Detection via Perception Pretext Reinforcement Learning}



  \icmlsetsymbol{equal}{*}
  \icmlsetsymbol{lead}{$^\dagger$}

  \begin{icmlauthorlist}
    \icmlauthor{Hao Tan}{a1,comp}
    \icmlauthor{Jun Lan}{lead,comp}
    \icmlauthor{Senyuan Shi}{a1}
    \icmlauthor{Zichang Tan}{a3}
    \icmlauthor{Zijian Yu}{comp}
    \icmlauthor{Huijia Zhu}{comp}
    \icmlauthor{Weiqiang Wang}{comp}\\
    \icmlauthor{Jun Wan}{a1,a2}
    \icmlauthor{Zhen Lei}{a1,a2}
  \end{icmlauthorlist}

  \icmlaffiliation{a1}{MAIS, Institute of Automation, Chinese Academy of Sciences}
  \icmlaffiliation{comp}{Ant Group}
  \icmlaffiliation{a2}{School of Artificial Intelligence, University of Chinese Academy of Sciences}
  \icmlaffiliation{a3}{Shenzhen Institute of Advanced Technology (SIAT), Chinese Academy of Sciences}
  
  \icmlcorrespondingauthor{Jun Wan}{jun.wan@ia.ac.cn}

  \begin{center}
    \faGithub\ \textbf{Project:} \url{https://github.com/EricTan7/VideoVeritas}
  \end{center}

  \icmlkeywords{Machine Learning, ICML}

  \vskip 0.3in
]



\printAffiliationsAndNotice{\icmlProjectLeader}

\begin{abstract}
  The growing capability of video generation poses escalating security risks, making reliable detection increasingly essential.
  In this paper, we introduce \videover, a framework that integrates fine-grained perception and fact-based reasoning.
  We observe that while current multi-modal large language models (MLLMs) exhibit strong reasoning capacity, their granular perception ability remains limited.
  To mitigate this, we introduce \textit{Joint Preference Alignment} and \textit{Perception Pretext Reinforcement Learning (PPRL)}.
  Specifically, rather than directly optimizing for detection task, we adopt general spatiotemporal grounding and self-supervised object counting in the RL stage, enhancing detection performance with simple \textit{perception pretext tasks}.
  To facilitate robust evaluation, we further introduce \textbf{MintVid}, a light yet high-quality dataset containing $3$K videos from $9$ state-of-the-art generators, along with a real-world collected subset that has factual errors in content.
  Experimental results demonstrate that existing methods tend to bias towards either \textit{superficial} reasoning or \textit{mechanical} analysis, while \videover$\,$ achieves more balanced performance across diverse benchmarks.
\end{abstract}

\section{Introduction}
\label{sec:intro}
With the rapid advancement of video generation, our digital lives have been immensely enriched.
AI‑generated videos have become pervasive on short-video platforms such as TikTok, which, while providing unprecedented entertainment, raise significant concerns regarding security.
AI‑generated video detection, which aims at determining the authenticity of a video, has therefore emerged as a critical topic.

\begin{figure}[t]
    \centering
    \includegraphics[width=0.96\linewidth]{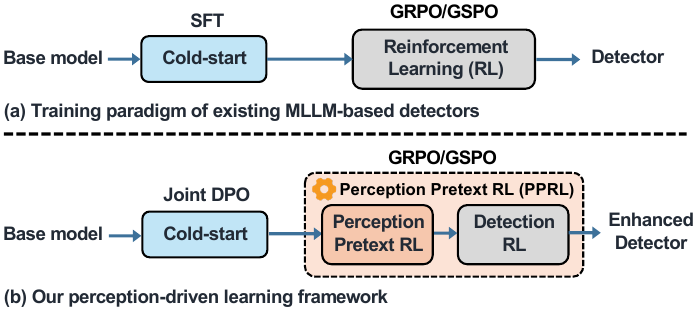}
     \caption{\textbf{Comparison with previous training pipeline.} (a) Existing MLLM-based detectors typically adopt supervised fine-tuning (SFT) or reinforcement learning (RL) on detection task. (b) Our framework adopts Joint DPO for cold-start, and further enhances the detection capacity by introducing simple perception pretext tasks in the RL stage.}
	\label{fig:intro}
    \vspace{-0.2cm}
\end{figure}

In this landscape, various methods~\citep{chen2024demamba, ma2025detecting, zheng2025d3, interno2025ai, zhang2025physics, corvi2025seeing} and datasets~\citep{chen2024demamba, ni2025genvidbench, wen2025busterx, wen2025busterx++, fu2025learning, li2025skyra} have been proposed.
Beyond binary discrimination, recent efforts~\citep{wen2025busterx++, park2025vidguard, fu2025learning, li2025skyra, gao2025david} focus on providing artifacts explanations.
However, current methods face \textbf{two challenges:}
\textbf{(1)} Even state-of-the-art (SoTA) multi-modal large language models (MLLMs) struggle to capture human-perceivable artifacts~\citep{wang2025video}.
To mitigate this, BusterX++~\citep{wen2025busterx++} adopts pure Reinforcement Learning (RL) on large-scale detection dataset, but the resulting model falls into superficial analysis, focusing on coarse features like environment and lighting.
\textbf{(2)} The task-oriented fine-tuning leads to \textit{mechanical grounded analysis}.
Methods like Skyra~\citep{li2025skyra} and DeepTraceReward~\citep{fu2025learning} construct grounded Chain-of-Thought (CoT) for Supervised Fine-Tuning (SFT), aiming to inject fine-grained perceptual capacities to base model.
However, the resulting model fails to conduct basic fact-based reasoning on AI parodies.

\begin{figure}[t]
    \centering
    \includegraphics[width=0.98\linewidth]{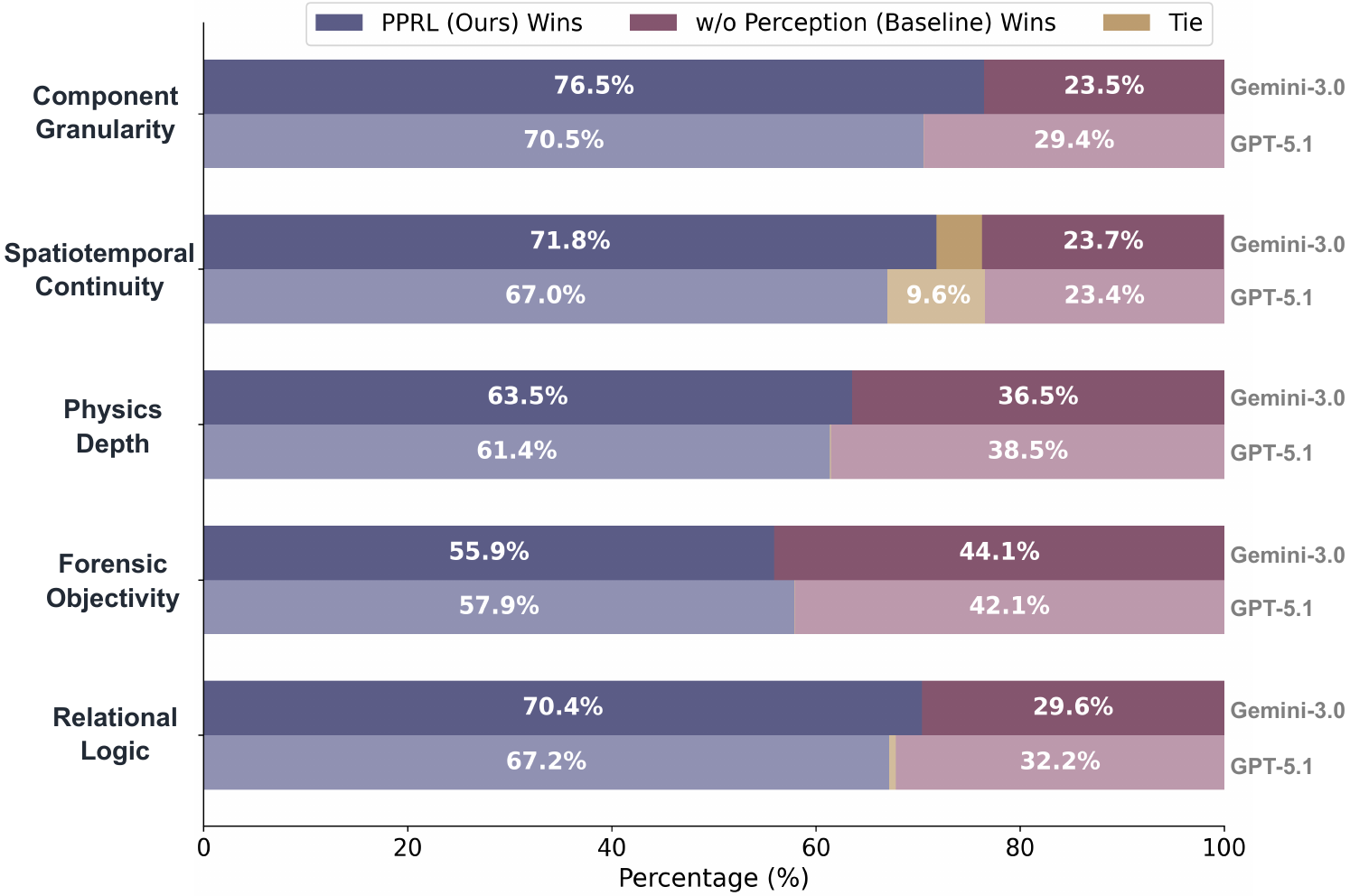}
     \caption{\textbf{To understand why PPRL enhances detection}, we characterize the model’s reasoning behavior across five distinct dimensions, finding that PPRL effectively shapes better reasoning behavior.
     For instance, model trained with PPRL tends to break down a whole scene into specific objects (i.e., ``Component Granularity'': 76.5\% win rate).
     Details are provided in Sec.~\ref{sec:further_ana}.}
	\label{fig:reason_ana}
    \vspace{-0.3cm}
\end{figure}

To mitigate these issues, we introduce \videover, an MLLM-based detector that integrates fine-grained perception and general reasoning.
As shown in Figure~\ref{fig:intro}, our approach leverages a two-stage training pipeline.
\textbf{In the first stage}, rather than performing SFT on large-scale database, we introduce Joint Preference Alignment.
Specifically, we construct question-answering (QA) reports to select data with diversified artifacts.
To integrate perceptual and reasoning abilities, we incorporate both response-level and video-level alignments, leveraging the base model itself as a strong guidance (e.g., using ``anti-label'' strategy to probe the intrinsic hallucinations as non-preference).
\textbf{In the second stage}, we introduce Perception Pretext RL (PPRL).
Rather than directly optimizing for AIGC detection task, we adopt (1) general spatiotemporal grounding and (2) self-supervised object counting as perception pretext tasks, which yield promising improvements on detection without requiring label-intensive annotations.
As shown in Figure~\ref{fig:reason_ana}, we also empirically demonstrate that PPRL leads to \textit{better reasoning behavior}, which therefore \textit{benefits} detection.

Moreover, videos in existing datasets (e.g., GenVideo~\cite{chen2024demamba} and GenVidBench~\cite{ni2025genvidbench}) primarily come from early models like ModelScope, which suffer from suboptimal temporal consistency.
To support more robust evaluation,
we present \textbf{MintVid}, a light AI-generated video dataset spanning three parts:
\textbf{(1)} $1.5$K highly realistic videos from $6$ powerful proprietary models, each including both T2V and TI2V,
\textbf{(2)} $2$K deepfake videos using $3$ SoTA public models,
and \textbf{(3)} a fact-based subset collected from short-video platforms, which could be identified through factual reasoning.
Together with available benchmarks, e.g., GenBuster++~\cite{wen2025busterx++}, the models are evaluated on a holistic and challenging setting.

Building on MintVid, we reveal that binary detectors like DeMamba~\cite{chen2024demamba} and RestraV~\cite{interno2025ai} struggle to yield satisfactory results.
MLLM with cold-start (e.g., Skyra) exhibits notably low recall on fact-based subset, while pure RL (e.g., BusterX++) demonstrates degraded results on those challenging subsets.
In contrast, \videover $\,$ achieves a more balanced performance.

To sum up, our main contributions are:
\begin{itemize}[left=8pt]
    \item We introduce Perception Pretext RL \textbf{algorithm}, which leverages simple perception pretext tasks to elevate detection performance. This method can be seamlessly integrated into existing R1-paradigm framework.
    \item We establish \videover, a \textbf{framework} that integrates fine-grained perception and fact-based reasoning for AI-generated video detection, achieving superior results over previous methods across multiple datasets.
    \item We introduce MintVid, a light yet high-quality \textbf{dataset} containing $3$K videos from $9$ SoTA generators. MintVid facilitates evaluations on three aspects, including general-content, facial and fact-based scenarios.
\end{itemize}

\begin{figure*}[t]
    \centering
    \includegraphics[width=0.98\linewidth]{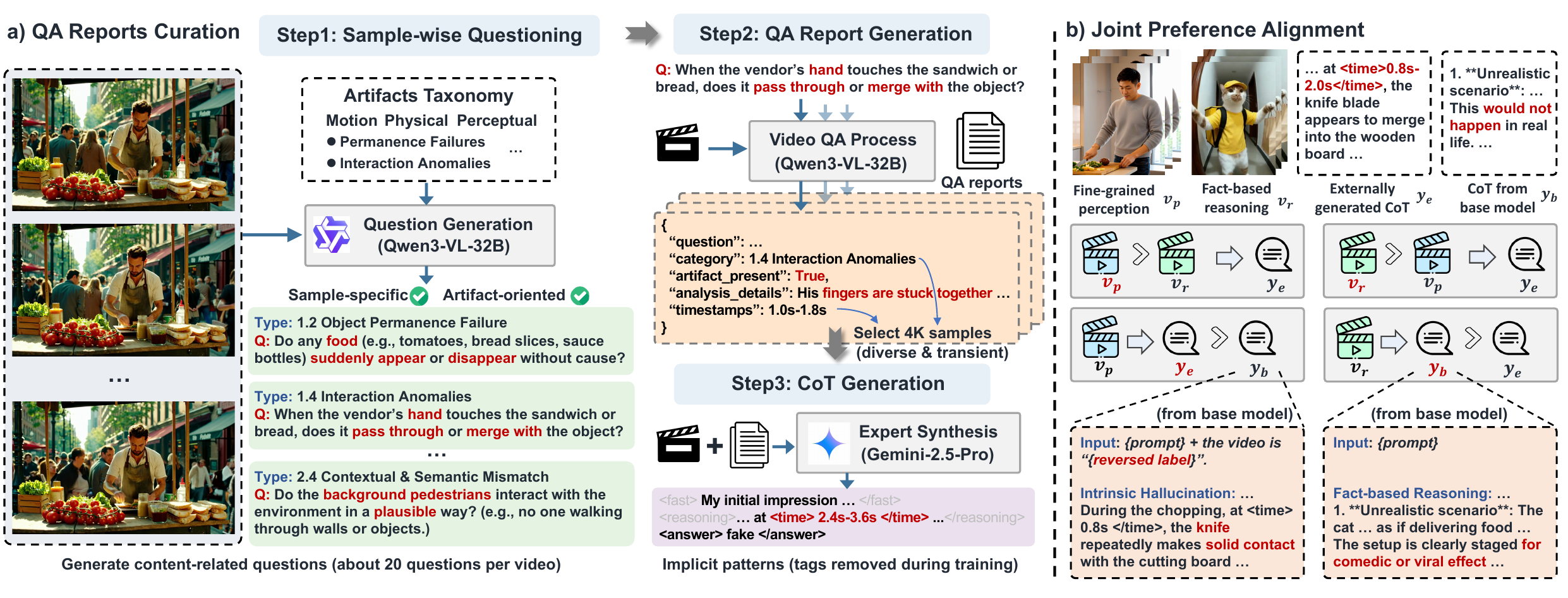}
     \caption{\textbf{Overview of our Joint Preference Alignment stage.} \textbf{(a)} We generate Question-Answering (QA) Reports to select diverse data and curate high-quality Chain-of-Thought (CoT). It involves generating artifact-oriented questions and creating detailed QA reports. \textbf{(b)} Joint DPO constructs preference pairs for both response-level and video-level alignments, leveraging external CoT and the base model's own reasoning to effectively guide the model.
     The artifacts taxonomy is provided in Appendix~\ref{sec:artifact}.}
	\label{fig:dpo}
\end{figure*}

\section{Related Work}
\subsection{AI-Generated Video Detection and Datasets}
Early studies primarily focused on image-level forgery detection~\citep{ojha2023towards, yan2024transcending, tan2024rethinking, nguyen2024laa,fu2025exploring, yan2024sanity, yan2024orthogonal, yang2025all}.
With the increasing capabilities of video generation, 
various efforts are made for video detection~\citep{chen2024demamba, bai2024ai, liu2024turns, ma2025detecting, zheng2025d3, interno2025ai, zhang2025physics, corvi2025seeing}.
For instance, D3~\citep{zheng2025d3} revealed the discrepancy in the second-order differences of features between real and AI-generated videos, enabling a training-free detection through simple pre-trained encoders.
Sharing similar insights, ReStraV~\citep{interno2025ai} leverages the statistical discrepancy between pre-trained representations for discrimination.
NSG-VD~\citep{zhang2025physics} introduces a novel perspective based on physical conservation principles.
Similar to recent ``bias-free'' approaches~\citep{zhou2024freqblender, yan2024transcending, kashiani2025freqdebias, chen2025dual} in image domain, WaveRep~\citep{corvi2025seeing} explores unbiased training using video VAE and frequency-level alignment, greatly enhancing detection robustness under H.264 compression.
Besides, some methods also target deepfake video detection~\citep{haliassos2021lips, wang2023altfreezing, xu2023tall, han2025towards, Kim_2025_ICCV, nguyen2025vulnerability, yan2025generalizing}, incorporating facial priors to achieve robust detection.
However, most methods are evaluated on previous datasets, e.g., GenVideo~\citep{chen2024demamba} and GenVidBench~\citep{ni2025genvidbench}, where most videos are derived from \textit{outdated} generative models, suffering from limited temporal consistency.
Although various benchmarks~\citep{du2025forensichub, wang2025forensics, huang2025so, li2025artificial} have been proposed recently, they primarily focus on image domain. 
In this paper, we introduce a video dataset to facilitate more robust evaluation.

\subsection{Multimodal Large Language Models}

\noindent \textbf{Brief Review of Generic MLLMs}.
To encode the temporal information, Qwen2-VL~\citep{wang2024qwen2} employs Multimodal Rotary Position Embedding (M‑RoPE). 
Qwen2.5‑VL~\citep{bai2025qwen2} further incorporates absolute temporal encoding, while Qwen3‑VL~\citep{bai2025qwen3} introduces explicit timestamp tokens, achieving strong performance on generic video benchmarks like VideoMME~\citep{fu2025video}.
To adaptively encode long videos, Keye‑VL1.5~\citep{yang2025kwai} proposes a slow–fast frames encoding strategy, while VideoLLaMA3~\citep{zhang2025videollama} introduces a Differential Frame Pruner, greatly improving performance on long video understanding.
Although their semantic‑level understanding capabilities have greatly advanced, their fine‑grained spatiotemporal perception remains limited~\citep{zhao2025videoperceiver}.

\noindent \textbf{MLLMs for AI-Generated Video Detection}.
Leveraging the semantic understanding capabilities of MLLMs, several studies have explored explainable AI‑generated video detection~\citep{song2024learning, jiang2025ivy, xu2025avatarshield, gao2025david, fu2025learning, park2025vidguard, wen2025busterx, wen2025busterx++, li2025skyra}.
DAVID-XR1~\citep{gao2025david} and VidGuard‑R1~\citep{park2025vidguard} employ Supervised Fine-Tuning (SFT) using Chain-of-Thought (CoT) generated by Gemini-2.5-Pro and Qwen2.5‑VL-72B, respectively.
DeeptraceReward~\citep{fu2025learning} introduces ``human-perceivable'' explainability and involves timestamps and bounding boxes in the reasoning process, while Skyra~\citep{li2025skyra} uses $4$K manually annotated samples to inject the grounding capacities into base model.
In contrast, BusterX++~\citep{wen2025busterx++} supposes that low‑quality cold starts can affect the generic reasoning ability and therefore adopts pure Reinforcement Learning (RL) on large-scale training data.
In this paper, we introduce Perception Pretext RL, which achieves promising gains without the burden of human-annotated AIGC data.

\begin{figure*}[t]
    \centering
    \includegraphics[width=0.98\linewidth]{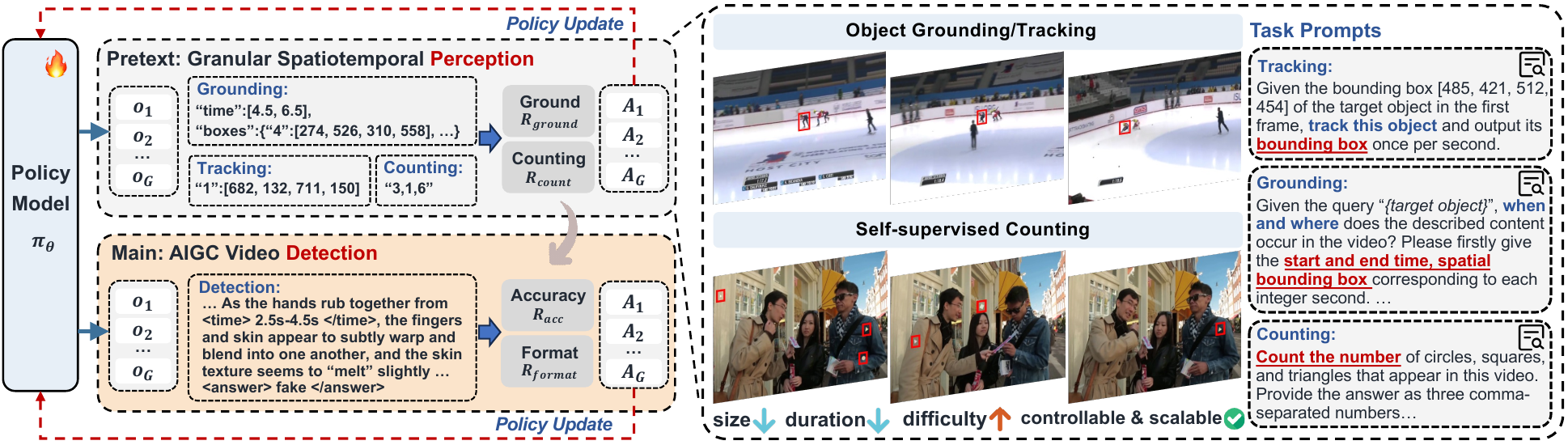}
     \caption{\textbf{Perception Pretext RL (PPRL)}. \textbf{Left}: \textit{Perception} is taken as a foundational phase to \textit{detection} . 
     The pretext phase can be implemented with various perception-oriented tasks, e.g., spatiotemporal grounding and object counting.
     \textbf{Right}: Examples of perception pretext tasks.
     Grounding and tracking data is sampled from OneThinker~\cite{feng2025onethinker}.
     The model is prompted to output exact bounding boxes and timestamps.
     Self-supervised counting is controllable by adjusting the size and duration of the objects, and the model is required to output the exact quantity of each shape.}
	\label{fig:stage2}
    \vspace{-0.2cm}
\end{figure*}

\section{Method}
In this section, we present a detailed description of the two-stage training pipeline and MintVid dataset.
\subsection{Joint Preference Alignment}
\label{sec:3.1}

\noindent \textbf{QA reports generation and CoT curation.}
As shown in Figure~\ref{fig:dpo}, for each video, we first generate various \textit{artifact-oriented} questions that are strongly \textit{related} to its content.
Then we use Qwen3-VL-32B~\cite{bai2025qwen3} to answer each question, forming a comprehensive QA report for each video.
Each item contains the type of artifacts and exact timestamps.
Such strategy offers two benefits:
(1) It enables \textit{independent} and label-agnostic perception of specific artifact, showing fewer hallucinations than directly prompted with Ground-Truth labels.
(2) It provides a valuable reference for data sampling.
Based on this pipeline, we filter $4$K samples from the database comprising GenBuster-200K~\cite{wen2025busterx}, RewardData~\cite{fu2025learning}, TalkingheadBench~\cite{xiong2025talkingheadbench} and several fact-based videos, where $\bm{v}_p$ and $\bm{v}_r$ denote videos that require fine-grained perception and fact-based reasoning, respectively.
We utilize Gemini-2.5-Pro to generate CoT based on its own reasoning and QA reports.
Following Veritas~\cite{tan2025veritas}, we adopt a structured reasoning design, but remove the explicit ``tags'' during training, encouraging the model to learn the underlying thinking pattern inherently.

\noindent \textbf{Response-level alignment.}
Suppose previous annotated CoT is denoted as $\bm{y}_e$,
we make full use of the base model itself for more effective alignments:
\textbf{(1)} We utilize ``anti-label'' strategy (i.e., insert the reversed label in the prompt) to trigger the base model's \textit{intrinsic} hallucinations, designating these as non-preferred responses.
\textbf{(2)} For videos requiring factual reasoning, the base model’s outputs are directly leveraged as preferred responses.
Let $\bm{y}_b$ denote the CoT generated from base model.
For $\bm{v}_p$ we assume that $\bm{y}_e$ is preferred over $\bm{y}_b$ (i.e., $\bm{y}_e\succ\bm{y}_b$), whereas for $\bm{v}_r$, we suppose the opposite preference (i.e., $\bm{y}_b\succ\bm{y}_e$).
Given the input query $\bm{x}$,
the response-level preference objective is formulated as:
\begin{equation}
\begin{aligned}
    \!u_t(\bm{x},\! \bm{v},\! \bm{y}_w,& \bm{y}_l\!) \!=\! \beta\!\log\!\frac{\pi_{\theta}(\bm{y}_w|\bm{v},\!\bm{x}\!)}{\pi_{{\text{ref}}}(\bm{y}_w\!|\bm{v},\!\bm{x}\!)}\!-\!\beta\!\log\!\frac{\pi_{\theta}(\bm{y}_l|\bm{v},\!\bm{x}\!)}{\pi_{{\text{ref}}}(\bm{y}_l\!|\bm{v},\!\bm{x}\!)}\!,\\
    \mathcal{L}_{\text{DPO}_t} = &-\mathbb{E}_{(\bm{x},\bm{v}_p,\bm{y}_e,\bm{y}_b)}\!
    \left[\log\sigma\! \left(u_t(\bm{x}, \bm{v}_p, \bm{y}_e, \bm{y}_b) \right)\right]\\
    &-\mathbb{E}_{(\bm{x},\bm{v}_r,\bm{y}_b,\bm{y}_e)}\!
    \left[\log\sigma\! \left(u_t(\bm{x}, \bm{v}_r, \bm{y}_b, \bm{y}_e) \right)\right]\!,
\end{aligned}
\end{equation}
where $\sigma(\cdot)$ denotes sigmoid function.

\noindent \textbf{Video-level alignments.}
To encourage the model to generate preferred outputs based on pure visual information, we involve video-level alignments.
Specifically, we encourage the model to output fine-grained analysis $\bm{y}_e$ based on video $\bm{v}_p$ (i.e., $\bm{v}_p\succ\bm{y}_r$), while outputing factual reasoning $\bm{y}_b$ when given video $\bm{v}_r$ (i.e., $\bm{v}_r\succ\bm{y}_p$).
The video-level preference objective is computed as:
\begin{equation}
\begin{aligned}
    \!u_v(\bm{x},\! \bm{y},\! \bm{v}_w,& \bm{v}_l\!) \!=\! \beta\!\log\!\frac{\pi_{\theta}(\bm{y}|\bm{v}_w,\!\bm{x}\!)}{\pi_{{\text{ref}}}(\bm{y}|\!\bm{v}_w,\!\bm{x}\!)}\!-\!\beta\!\log\!\!\frac{\pi_{\theta}(\bm{y}|\bm{v}_l,\!\bm{x}\!)}{\pi_{{\text{ref}}}(\bm{y}|\!\bm{v}_l,\!\bm{x}\!)}\!,\\
    \!\mathcal{L}_{\text{DPO}_v} = &-\mathbb{E}_{(\bm{x},\bm{y}_e,\bm{v}_p,\bm{v}_r)}\!
    \left[\log\sigma\! \left(u_v(\bm{x}, \bm{y}_e,\! \bm{v}_p,\! \bm{v}_r) \right)\right]\\
    &-\mathbb{E}_{(\bm{x},\bm{y}_b,\bm{v}_r,\bm{v}_p)}\!
    \left[\log\sigma\! \left(u_v(\bm{x}, \bm{y}_b,\! \bm{v}_r,\! \bm{v}_p) \right)\right]\!.
\end{aligned}
\end{equation}
The final objective for the Joint DPO is defined as:
\begin{equation}
    \mathcal{L}_{\text{J-DPO}}=\mathcal{L}_{\text{DPO}_t} + \mathcal{L}_{\text{DPO}_v}.
\end{equation}

\subsection{Perception Pretext RL (PPRL)
}
\label{sec:rl}

As illustrated in Figure~\ref{fig:stage2}, 
rather than directly optimizing for AIGC detection task, 
we introduce perception pretext tasks.
By first encouraging the model to accurately perceive subtle and transient targets,
the fine-grained perception capacity is sharpened, benefiting the following detection task.
Specifically, we introduce two types of perception tasks: (1) general spatiotemporal grounding and (2) self-supervised object counting.

\noindent \textbf{General grounding.}
This includes spatiotemporal grounding and object tracking.
For tracking task, given the bounding box $B_1$ in the first frame, the model is required to predict a sequence of bounding boxes $\{\widehat{B}_i\}_{i=2}^N$ to track the object across the video.
Suppose $\{B_i\}_{i=2}^N$ denotes the ground-truth boxes,
the reward is measured as the mean Intersection over Union (IoU) over all frames:
\begin{equation}
    R_{\text{track}} = \sum_{i=2}^N \frac{|\widehat{B}_i \cap B_i|}{|\widehat{B}_i \cup B_i|}.
\end{equation}
For spatiotemporal grounding, the model is required to predict both the temporal span $\widehat{t}$ of a query and its bounding boxes $\{{\widehat{B}_i}\}_{i=1}^M$ across the video.
Given the ground-truth $t_y$ and $\{B_i\}_{i=1}^M$,
the reward is measured according to both the timestamp and spatial boxes:
\begin{equation}
    R_{\text{ground}} = \frac{1}{2}\frac{|\widehat{t} \cap t_y|}{|\widehat{t} \cup t_y|} + \frac{1}{2} \sum_{i=1}^M \frac{|\widehat{B}_i \cap B_i|}{|\widehat{B}_i \cup B_i|}.
\end{equation}
There are abundant samples that are publicly available.
In our case, all data are sampled from OneThinker~\cite{feng2025onethinker}, with easy samples filtered out, retaining only small and short-duration targets for training.

\noindent \textbf{Self-supervised object counting.}
For a given video, we randomly add a set of objects with randomized attributes (e.g., size, duration and position) to the frames.
The model is required to output the counts for each shape category in the video, including circles, squares and triangles.
For each shape $s$, the reward $R_{\text{count},s}$ is inversely proportional to the relative prediction error:
\begin{equation}
    R_{\text{count},s} = \text{max}(0, 1-\frac{|\widehat{y}_s-y_s|}{y_s+\epsilon}),
\end{equation}
where $\widehat{y}_s$ and $y_s$ denotes the predicted and ground-truth counts, respectively. $\epsilon$ is a small constant (e.g., $10^{-6}$) to ensure numerical stability.
The final counting reward $R_{\text{count}}$ is averaged from all shapes:
\begin{equation}
    R_{\text{count}} = \frac{1}{S}\sum_{s=1}^{S}R_{\text{count},s}.
\end{equation}
Since the object size and duration are fully controllable, we can generate diverse and challenging training signals at scale.
In our case, we randomly select videos from OpenVid-1M~\cite{nan2024openvid} to construct learning samples.
Only with precise spatiotemporal perception can the model correctly count different shapes.
This enables efficient scaling without requiring any manual annotations.

\noindent \textbf{AIGC detection.}
For the detection task, we utilize accuracy reward $R_{\text{acc}}$ and format reward $R_{\text{format}}$, where $R_{\text{acc}}=1$ when the answer is correct and $R_{\text{acc}}=0$ otherwise.
$R_{\text{format}}$ is given when the answer is enclosed in \texttt{<answer>} \texttt{</answer>} tags.
The reward for detection task is:
\begin{equation}
    R_{\text{detection}} = R_{\text{acc}} + \alpha R_{\text{format}},
\end{equation}
where $\alpha= 0.2$ is to control the effect of format reward.
Specifically, we use $3$K samples for general grounding, $2$K samples for self-supervised object counting, and $10$K samples (from database in Sec.~\ref{sec:3.1}) for AIGC detection.
The perception task and detection task are performed in a sequential manner.
Since our method is suitable for most R1-style algorithms~\cite{liu2024deepseek}, we adopt Group Sequence Policy Optimization (GSPO)~\cite{zheng2025group} for training.
Instead of relying on manually annotated AIGC data or elaborate reward design, we demonstrate that simple RL on perception pretext tasks could achieve promising gains.

\subsection{MintVid Dataset}
\label{sec:dataset}

\begin{figure}[t]
    \centering
    \includegraphics[width=0.98\linewidth]{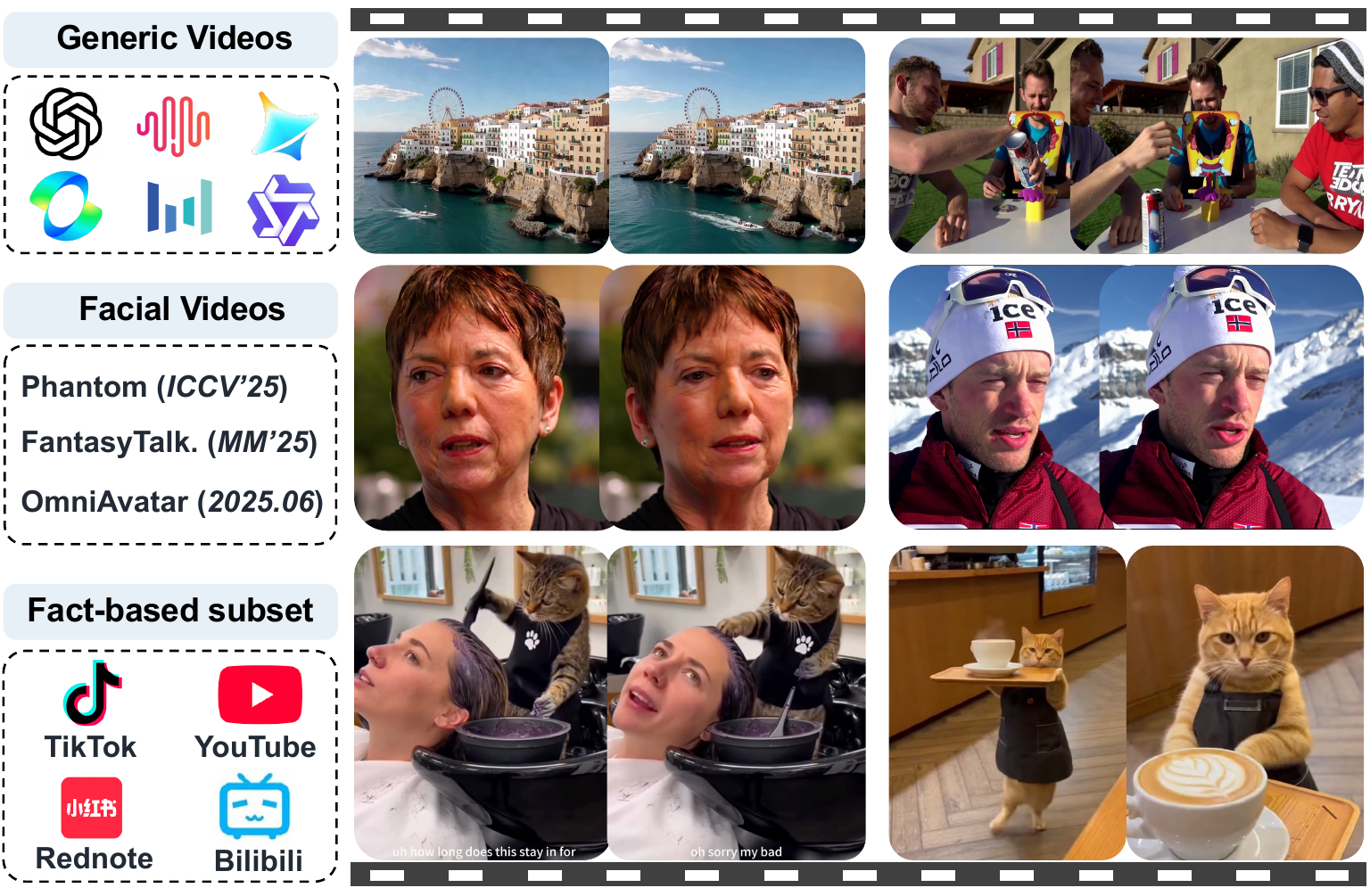}
     \caption{\textbf{Examples of the MintVid dataset.} MintVid comprises three different parts, which facilitates more complete evaluation.}
	\label{fig:mintvid}
    \vspace{-0.2cm}
\end{figure}

As shown in Figure~\ref{fig:mintvid}, 
MintVid comprises three parts:

\noindent \textbf{General content videos.} 
$1.5$K videos from $6$ powerful proprietary models: Jimeng 3.0 Pro, Seedance 1.0 Pro, Kling 2.5-Turbo, Wan2.5, MiniMax Hailuo 2.3, and Sora2.
For each model, we sample $100$ meticulous prompts from VideoVerse~\cite{wang2025videoverse}, PhyGenBench~\cite{meng2024towards} and VMBench~\cite{ling2025vmbench}, and $200$ prompts from OpenVid-1M~\cite{nan2024openvid} captions.
Among them, $100$ prompts are paired with the corresponding first frame to generate TI2V data.
After manual filtering, $1.5$K videos are kept.
Real videos are sampled from OpenVid-1M.

\noindent \textbf{Facial videos.} 
$2$K videos from $3$ SoTA human-centric video generative models: OmniAvatar~\cite{gan2025omniavatar}, FantasyTalking~\cite{wang2025fantasytalking}, and Phantom~\cite{liu2025phantom}, all using their 14B variants.
Real videos are sampled from VFHQ~\cite{xie2022vfhq} and HDTF~\cite{zhang2021flow}.
We curate input prompts to yield high-quality videos.
Please see Appendix~\ref{sec:app_dataset} for more details.

\noindent \textbf{Fact-based subset.}
This part is often overlooked in existing datasets.
For videos that violate objective facts, general MLLMs can perform correct reasoning, whereas fine-tuned models may fail due to formulaic analysis.
To this end, we collect over $200$ videos (including both real and fake) from TikTok, YouTube, and Bilibili, and conduct manual filtering to ensure all videos can be verified using objective facts.

\begin{table*}[t]
\caption{Performance comparison on video datasets. Accuracy is reported except for D3~\cite{zheng2025d3}, where $*$ means Average Precision (AP) is adopted following the official guideline. 
The final average performance is calculated by first averaging the results across ID (in-domain), OOD (out-of-domain) and OOD-MintVid, and then taking the average of the three values.
The best results are \textbf{bolded} and the second best are \underline{underlined}.
More metrics are provided in Appendix~\ref{sec:more_res}.}
\vspace{-3pt}
    \centering
    \renewcommand\arraystretch{1.1}
    \scalebox{0.82}{
        \small
        \begin{tabular}{p{86pt}<{\raggedright}p{15pt}<{\centering}p{15pt}<{\centering}p{15pt}<{\centering}p{15pt}<{\centering}p{15pt}<{\centering}p{15pt}<{\centering}p{15pt}<{\centering}p{15pt}<{\centering}p{15pt}<{\centering}p{15pt}<{\centering}p{15pt}<{\centering}p{15pt}<{\centering}p{15pt}<{\centering}p{15pt}<{\centering}p{15pt}<{\centering}p{15pt}<{\centering}p{20pt}<{\centering}p{16pt}<{\centering}}
        \toprule[1pt]
        \multirow{2}{*}{\hspace{-5pt}\vspace{-2.0em}\textbf{Method}} & \multirow{2}{*}{\vspace{-2.0em}\textbf{ID}} & \multicolumn{6}{c}{\textbf{OOD}} & \multicolumn{10}{c}{\textbf{OOD-MintVid}} & \multirow{2}{*}{\vspace{-2.0em}\textbf{Avg.}} \\
        \cmidrule(lr){3-8} \cmidrule(lr){9-18}
        & & \rotatebox{40}{GenBuster++} & \rotatebox{40}{LOKI} & \rotatebox{40}{Vidu Q1} & \rotatebox{40}{Gen-4} & \rotatebox{40}{Veo3} & \rotatebox{40}{Emu3} & \rotatebox{40}{Phantom-14B} & \rotatebox{40}{OmniAvatar} & \rotatebox{40}{FantasyTalking} & \rotatebox{40}{Seedance1.0Pro} & \rotatebox{40}{Jimeng3.0Pro} & \rotatebox{40}{Kling2.5-Turbo} & \rotatebox{40}{Hailuo2.3} & \rotatebox{40}{Wan2.5} & \rotatebox{40}{Sora2} & \rotatebox{40}{Fact} & \\
        \shline
        \rowcolor{lightblue}\textit{\textbf{Small Vision Models}} &&&&&&&&&&&&&&&&&&\\
        \hspace{-5pt}\rule{0pt}{7pt}DeMamba & \underline{87.6}  & \underline{86.8}  & 71.1  & 80.7  & 83.4  & 79.4  & 85.9  & 57.4  & 55.2 & \underline{62.2}  & 49.2  & 46.2  & 46.6  & 41.1  & 49.3  & 41.4  & 73.3  & 73.7 \\
        \hspace{-5pt}\rule{0pt}{7pt}D3$^*$ (\textit{ICCV'25}) & 49.5  & 46.8  & 39.1  & 55.1  & 89.0  & 66.1  & 90.0  & 54.8  & 36.4  & 44.1  & 61.6  & 48.7  & 51.7  & 78.9  & 50.5  & 54.6  & 32.4  & 55.1 \\
        \hspace{-5pt}\rule{0pt}{7pt}NSG-VD (\textit{NIPS'25}) & 52.4  & 53.3  & 55.1  & 53.4  & 58.2  & 64.3  & 50.7  & 54.3  & 53.8  & 58.5  & 53.8  & 58.0  & 56.0  & 69.4  & 57.0  & 57.8  & 52.8  & 55.1 \\
        \hspace{-5pt}\rule{0pt}{7pt}ReStraV (\textit{NIPS'25}) & 50.7  & 48.9  & 64.8  & 54.8  & 52.8  & 59.8  & 95.2  & 49.6  & 49.3  & 49.9  & 36.9  & 48.4  & 47.2  & 30.2  & 40.9  & 45.4  & 65.0  & 53.2 \\
        \shline
        \rowcolor{lightblue}\textit{\textbf{Generic MLLMs}} &&&&&&&&&&&&&&&&&&\\
        \hspace{-5pt}Qwen2.5-VL-7B & 54.1  & 59.5  & 49.4  & 58.8  & 53.3  & 60.3  & 99.2  & 50.7  & 50.0  & 49.7  & 63.6  & 49.4  & 51.6  & 53.1  & 55.8  & 50.4  & 87.1  & 57.9 \\
        \hspace{-5pt}Qwen3-VL-8B & 65.1  & 62.4  & 56.6  & 60.2  & 59.4  & 62.9  & \textbf{99.9}  & 53.3  & 50.6  & 50.3  & 64.2  & 51.8  & 55.1  & 54.1  & 66.7  & 50.8  & 91.9  & 63.6   \\
        \hspace{-5pt}InternVL3.5-8B & 51.7  & 59.6  & 44.1  & 56.1  & 53.4  & 56.7  & \underline{99.7}  & 51.3  & 50.2  & 49.5  & 62.6  & 49.8  & 49.1  & 50.2  & 53.9  & 54.1  & 91.7  & 56.5  \\
        \hspace{-5pt}MiMo-VL-7B & 65.8  & 54.3  & 53.6  & 65.5  & 65.8  & 66.0  & 98.8  & 58.4  & 50.1  & 50.6  & 69.7  & 56.7  & 61.7  & 56.1  & 66.0  & 55.4  & 89.9  & 64.8  \\
        \hspace{-5pt}Keye-VL-1.5-8B & 52.8  & 62.4  & 60.4  & 58.3  & 56.0  & 62.4  & 94.7  & 51.3  & 51.1  & 53.3  & 59.7  & 50.5  & 51.0  & 52.8  & 52.9  & 51.8  & 80.4  & 58.0  \\
        \hspace{-5pt}GLM-4.5V & 61.7  & 59.4  & 49.4  & 56.5  & 56.6  & 62.4  & 97.1  & 57.9  & 52.2  & 50.5  & 63.2  & 48.9  & 49.2  & 49.8  & 58.3  & 47.9  & 88.2  & 60.6 \\
        \hspace{-5pt}Qwen3-VL-235B-A22B & 66.7  & 64.6  & 57.9  & 57.4  & 65.7  & 63.4  & 98.5  & 54.1  & 50.9  & 51.2  & 65.8  & 49.1  & 32.3  & 49.8  & 63.5  & 50.5  & \underline{93.6}  & 63.6  \\
        \hspace{-5pt}Gemini-2.5-Pro & 80.5  & 71.8  & 70.2  & 70.8  & 76.0  & 70.4  & 93.3  & 57.8  & 55.8  & 55.6  & 69.4  & 60.2  & 65.6  & 62.7  & 65.7  & \underline{60.9}  & 88.3  & 73.4 \\
        \hspace{-5pt}Gemini-3-Pro-Preview & 79.4  & 82.5  & \underline{74.9}  & \underline{83.1}  & 76.3  & \underline{85.7}  & 89.3  & \underline{65.5}  & \textbf{65.3}  & 61.5  & \underline{81.4}  & \underline{72.1}  & 73.1  & \underline{74.3}  & 73.1  & 63.5  & 91.4  & \underline{77.8} \\
        \shline
        \rowcolor{lightblue}\multicolumn{4}{c}{\hspace{-22pt}\textit{\textbf{MLLM-based Video Forgery Detectors}}} &&&&&&&&&&&&&&&\\
        \hspace{-5pt}BusterX++ & 77.1  & 79.6  & 70.9  & 53.0  & \underline{89.9}  & 76.2  & 98.9  & 62.3  & 48.6  & 47.6  & 71.8  & 71.7  & \underline{82.3}  & 70.4  & \textbf{87.3}  & 58.7  & 88.8  & 74.7  \\
        \hspace{-5pt}Skyra-RL & 52.1  & 55.7  & 37.7  & 52.9  & 59.1  & 55.9  & 51.5  & 50.1  & 49.9  & 49.9  & 63.4  & 51.7  & 53.5  & 51.7  & 59.6  & 50.6  & 51.9  & 52.5  \\
        \rowcolor{Gray}\hspace{-7pt} \videover $\,$(\textbf{ours}) & \textbf{93.1}  & \textbf{91.4}  & \textbf{78.1}  & \textbf{93.6}  & \textbf{96.7}  & \textbf{94.5}  & 99.4  & \textbf{79.0}  & \underline{56.2}  & \textbf{84.5}  & \textbf{86.6}  & \textbf{80.8}  & \textbf{86.0}  & \textbf{85.8}  & \underline{86.3}  & \textbf{67.6}  & \textbf{96.1}  & \textbf{88.8} \\
        \bottomrule[1pt]
        \end{tabular}
    }
    \label{tab:main}
    \vspace{-0.2cm}
\end{table*}

\section{Experiments}
\subsection{Experimental Setup}

\noindent \textbf{Evaluation Protocols.}
In-domain (ID) testing includes test sets of GenBuster-200K~\cite{wen2025busterx} and TalkingheadBench~\cite{xiong2025talkingheadbench}.
Out-of-domain (OOD) testing involves existing benchmarks GenBuster++~\cite{wen2025busterx++}, LOKI~\cite{ye2024loki} (using their video set), and high-quality subsets (e.g., Gen-4) from VBench-2~\cite{zheng2025vbench} and MVAD~\cite{hu2025mvad}.
MintVid is adopted as a challenging OOD testing.
$4$ SoTA binary detectors are trained on our dataset using their original preprocessing.
Both public and proprietary MLLMs are evaluated using CoT prompting.
We also evaluate $2$ recent MLLM-based video forgery detectors, including BusterX++~\cite{wen2025busterx++} and Skyra-RL~\cite{li2025skyra}.

\noindent \textbf{Implementation Details.}
We implement the \videover$\,$ with Qwen3-VL-8B~\cite{bai2025qwen3}.
For Joint DPO, we train the model for $3$ epochs using LoRA~\cite{hu2022lora} (rank=$64$, $\alpha$=$128$), with a learning rate of $5\times 10^{-5}$ and batch size of $64$.
For PPRL, the model is trained for $2$ epochs with the same LoRA setting, and the learning rate is set to $1\times 10^{-5}$ with a batch size of $32$.
The temperature is set to $1.0$.
Each epoch contains both perception and detection tasks.
The videos are sampled at $3$ FPS for up to $5$ seconds, keeping their native resolutions.

\subsection{Main Results}

\noindent \textbf{Comparisons to binary detectors.}
As shown in Table~\ref{tab:main},
recent statistical-based methods~\cite{zheng2025d3, interno2025ai} are effective on certain subsets (e.g., 90.0\% and 95.2\% on Emu3), but their performance collapses to near-chance levels on most contemporary benchmarks, and even fails on in-domain subsets.
This suggests that statistical discrepancies may not be a robust criterion for more realistic videos.
DeMamba~\cite{chen2024demamba} performs well on previous datasets like GenBuster++, but suffers a sharp performance drop on the more challenging MintVid dataset.
In contrast, \videover$\,$ exhibits certain advantages, achieving 15.1\% averaged gains over the previous best.

\noindent \textbf{Comparisons to SOTA MLLMs.}
As shown in Table~\ref{tab:main}, although generic MLLMs demonstrate strong performance on the fact-based subset, their performance on other datasets remains limited, exhibiting low recall (presented in Table~\ref{tab:main_metrics}).
Notably, Gemini-3.0-Pro-Preview outperforms most open-source models and exhibits a \textit{clear lead} on MintVid.
Its average performance even surpasses finetuned models like DeMamba and BusterX++, indicating that \textit{a substantial gap still exists} between public and proprietary models.
Moreover, our approach achieves an average improvement of 25.2\% over base model, and it also outperforms those MLLMs with similar parameter scale, demonstrating the effectiveness of our training strategy.

\begin{figure*}[t]
    \scriptsize
    \centering
    \begin{minipage}{0.45\linewidth}
            \captionof{table}{Ablations on the type of perception pretext tasks. $5$K perception data are taken as default setting.}
    \label{tab:abl_task}
    \centering
    \renewcommand\arraystretch{1.07}
    \scalebox{0.82}{
        \small
        \begin{tabular}{p{20pt}<{\centering}p{20pt}<{\centering}p{20pt}<{\centering}p{15pt}<{\centering}p{15pt}<{\centering}p{15pt}<{\centering}p{15pt}<{\centering}p{15pt}<{\centering}p{18pt}<{\centering}}
       \multicolumn{3}{c}{\textbf{Pretext Tasks}} & \multicolumn{2}{c}{\textbf{ID}} & \multicolumn{2}{c}{\textbf{OOD}} & \multicolumn{2}{c}{\textbf{MintVid}} \\
       \cmidrule(lr){1-3} \cmidrule(lr){4-5} \cmidrule(lr){6-7} \cmidrule(lr){8-9}
       \textbf{SSL} & \textbf{G-G} & \textbf{A-G} & Acc & F1 & Acc & F1 & Acc & F1 \\
        \shline
        \rule{0pt}{7pt}\ding{55} & \ding{55} & \ding{55} & 92.9  & 93.4  & 91.2  & 92.1  & 77.8  & 73.3 \\
        \checkmark & \ding{55} & \ding{55} & 91.4  & 92.1  & 91.3  & 92.3  & \underline{80.4}  & \textbf{78.0}  \\
        \ding{55} & \checkmark & \ding{55} & 92.3  & 92.8  & 91.1  & 92.2  & 79.8  & 76.7 \\
        \ding{55} & \ding{55} & \checkmark & 92.7  & 93.2  & 91.2  & 92.4  & 79.5  & 76.9 \\
        \rowcolor{Gray}\checkmark & \checkmark & \ding{55} & \underline{93.1}  & \underline{93.5}  & \textbf{92.3}  & \textbf{93.4}  & \textbf{80.9}  & \underline{77.3} \\
        \checkmark & \ding{55} & \checkmark & 92.9  & 93.3  & \underline{92.0}  & \underline{92.9}  & 79.8  & 77.0 \\
        \ding{55} & \checkmark & \checkmark & \textbf{93.5}  & \textbf{93.9}  & 91.3  & 92.2  & 79.4  & 76.0 \\
        \checkmark & \checkmark & \checkmark & 92.7  & 93.2  & 91.3  & 92.2  & 78.6  & 75.0 \\
        \end{tabular}
    }
	\end{minipage}
    \hfill
    \begin{minipage}{0.52\linewidth}
            \includegraphics[height=4.0cm, width=0.99\linewidth]{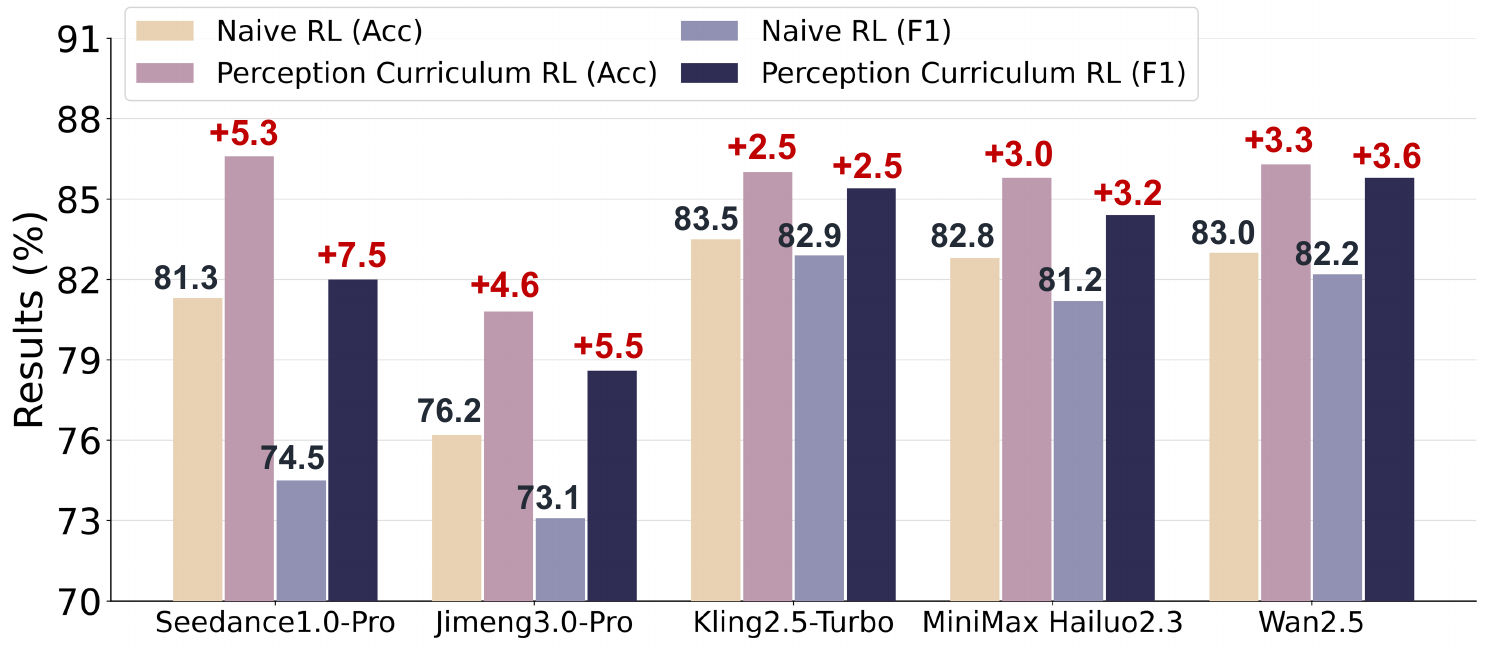}
    \vspace{-5pt}
    \caption{Improvements of the proposed PPRL. ``SSL+G-G'' is adopted in our method.}
	\label{fig:improve_perception}
	\end{minipage}
    \vspace{-0.3cm}
\end{figure*}

\noindent \textbf{Comparisons to MLLM-based detectors.}
As shown in Table~\ref{tab:main}, BusterX++ achieves 74.7\% average accuracy, but shows suboptimal performance on MintVid, especially the facial subsets.
Skyra-RL proves largely ineffective across most datasets.
We suppose this may stem from the strict preprocessing applied to its training and testing data (e.g., 256p resizing), which could bias the model toward exploiting minor inconsistencies.
In contrast, \videover$\,$ outperforms them on (1) existing datasets like GenBuster++ (+11.8\%) and LOKI (+7.2\%), (2) perception-heavy datasets like Jimeng3.0-Pro (+9.1\%), and (3) fact-dependent subset,
achieving a more balanced and robust detection capability.

\subsection{Ablation Studies}
\label{sec:ablation}

\begin{table}[t]
\caption{Ablations on the training stages.
$\dagger$ means using format reward~\cite{li2025skyra} to incentivize grounded reasoning.}
    \centering
    \renewcommand\arraystretch{1.06}
    \scalebox{0.89}{
        \small
        \begin{tabular}{p{60pt}<{\raggedright}p{20pt}<{\centering}p{20pt}<{\centering}p{20pt}<{\centering}p{20pt}<{\centering}}
        % \toprule[1pt]
        \multirow{2}{*}{\hspace{-5pt}\textbf{Method}} & \multirow{2}{*}{\textbf{ID}} & \multirow{2}{*}{\makebox[0pt][l]{\hspace{-1.0em}\textbf{OOD}}} & \multicolumn{2}{c}{\textbf{MintVid}} \\
        \cmidrule(lr){4-5}
        &&& Others & Fact \\
        \shline
        \hspace{-5pt}Base & 65.1 & 66.9 & 55.2 & 91.9 \\
        \hspace{-5pt}$+$ SFT & \textbf{91.9} & 82.5 & 68.3 & 90.7 \\
        \rowcolor{Gray}\hspace{-5pt}$+$ J-DPO & 91.4 & \textbf{83.0} & \textbf{69.1} & \textbf{94.6} \\
        $-$ DPO$_v$ & 91.3 & 82.7 & \textbf{69.1} & 92.9 \\
        $-$ DPO$_t$ & 89.2 & 80.9 & 66.3 & 88.3 \\
        \hline
        \hspace{-5pt}\rule{0pt}{8pt}$+$ Pure RL & 94.2 & 88.3 & 69.1 & \textbf{96.1} \\
        \hspace{-5pt}$+$ Pure RL$^\dagger$ & \textbf{94.5} & 90.2 & 70.9 & 87.5 \\
        \rowcolor{Gray}\hspace{-5pt}$+$ J-DPO $+$ RL & 93.1 & \textbf{92.3} & \textbf{79.2} & \textbf{96.1} \\
        % \bottomrule[1pt]
        \end{tabular}
    }
    \label{tab:abl_dpo}
    \vspace{-0.5cm}
\end{table}

\noindent \textbf{Ablations on the training stages.}
As shown in Table~\ref{tab:abl_dpo}, SFT and Joint DPO (J-DPO) is trained on $4$K samples. 
``J-DPO+RL'' is further trained on $10$K AIGC data and $5$K perception data.
Pure RL is trained on $20$K samples, which ensures similar data scale to investigate the effect of cold-start.
Specifically, we draw the following observations:
\textbf{(1)} Supervised Fine-Tuning (SFT) achieves great improvements (e.g., +13.1\% on MintVid-Others) but degrades the performance on Fact subset.
In contrast, J-DPO effectively \textit{integrates} the overall capacities, boosting accuracy on Fact subset to 94.6\%.
\textbf{(2)} Video-level alignment (DPO$_v$) is beneficial to synergizing the fact-based reasoning ability, while response-level alignment (DPO$_t$) is critical to overall performance.
\textbf{(3)} The performance of pure RL on challenging subsets remains limited (e.g., 69.1\% vs. 79.2\% compared to ours), indicating the necessity of cold-start.

\noindent \textbf{Ablations on PPRL: the type of perception tasks.}
Generally, the perception tasks can be categorized into three types with increasing annotation cost:
\textbf{(1) Self-Supervised Learning (SSL)}, where we use object counting as a zero-cost pretext for foundational perception. 
\textbf{(2) Generic Grounding (G-G)}, where abundant public data can be utilized.
\textbf{(3) Artifact Grounding (A-G)}, which is the most specialized and costly task.
As shown in Table~\ref{tab:abl_task}, each perception task independently improves the performance. 
Combining SSL and G-G yields further enhancements, achieving substantial gains on MintVid dataset as shown in Figure~\ref{fig:improve_perception},
suggesting that the \textit{perceptual acuity gained from these tasks} provides a better foundation for detection.
Interestingly, further combining A-G degrades the performance, which we suppose is due to the quality of A-G annotations.
For A-G, we use data sampled from Molmo2~\cite{clark2026molmo2}, which is coarse-grained (e.g., the entire person is masked out even only the arm deforms), making them less compatible with the fine-grained signals from SSL and G-G.
We suppose the results might be different if \textit{higher-quality artifact grounding annotations} are available. 
Overall, without any domain-specific annotations, using SSL and G-G as perceptual pretext tasks yields promising improvements.
Detailed results can be found in Appendix Table~\ref{tab:task_mintvid}.

\begin{figure}[t]
    \centering
    \includegraphics[width=0.96\linewidth]{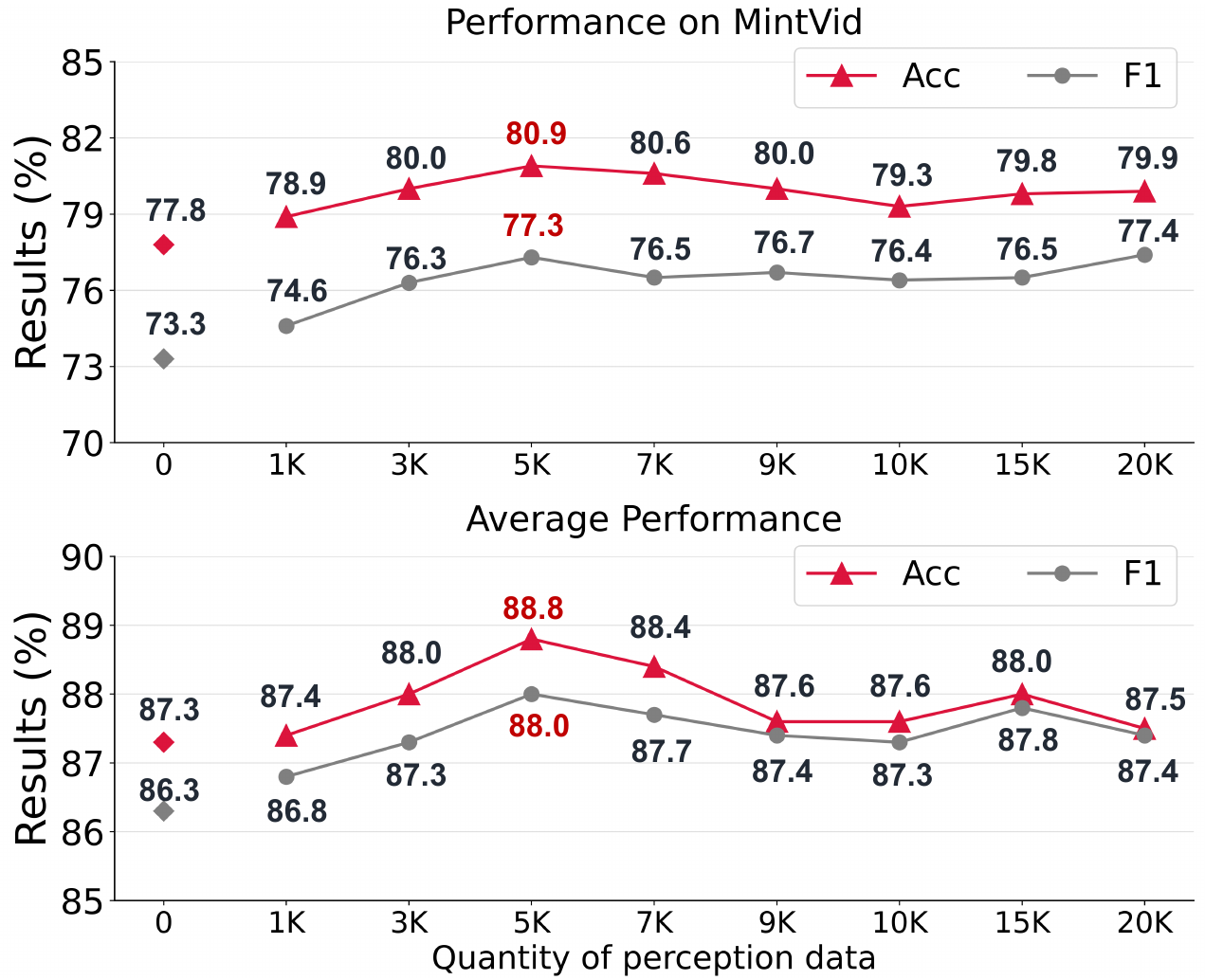}
     \caption{Ablations on the quantity of perception data. Accuracy and F1 are reported on MintVid (upper) and Average (lower). Quantity of ``0'' denotes the baseline.}
	\label{fig:data_ratio}
    \vspace{-0.3cm}
\end{figure}

\begin{figure*}[t]
    \scriptsize
    \centering
    \begin{minipage}{0.57\linewidth}
            \captionof{table}{Difficulty of the SSL task. ($\cdot$,$\cdot$) denotes the range of values (``\textit{the number of pixels per side}'' for ``Size'' and ``\textit{seconds}'' for ``Duration''). ``Baseline'' is the model trained exclusively on AIGC detection task.}
            \vspace{-5pt}
    \label{tab:abl_task}
    \centering
    \renewcommand\arraystretch{1.07}
    \scalebox{0.82}{
        \small
        \begin{tabular}{p{42pt}<{\raggedright}p{38pt}<{\centering}p{35pt}<{\centering}p{16pt}<{\centering}p{16pt}<{\centering}p{16pt}<{\centering}p{16pt}<{\centering}p{16pt}<{\centering}p{20pt}<{\centering}}
        \multirow{2}{*}{\hspace{-5pt}\textbf{Difficulty}} & \multirow{2}{*}{\textbf{Size}} & \multirow{2}{*}{\makebox[0pt][l]{\hspace{-2.0em}\textbf{Duration}}} & \multicolumn{2}{c}{\textbf{ID}} & \multicolumn{2}{c}{\textbf{OOD}} & \multicolumn{2}{c}{\textbf{MintVid}} \\
        \cmidrule(lr){4-5} \cmidrule(lr){6-7} \cmidrule(lr){8-9}
        &&& Acc & F1 & Acc & F1 & Acc & F1 \\
        \shline
        \hspace{-5pt}\gc{Baseline} & \gc{-} & \gc{-} & \underline{92.9}  & \underline{93.4}  & 91.2  & 92.1  & 77.8  & 73.3 \\
        \hspace{-5pt}Easy & (120, 240) & (3s, 5s) & 92.5  & 92.9  & 91.3  & 92.2  & 79.5  & 76.3  \\
        \hspace{-5pt}Medium & (40, 120) & (2s, 4s) & \underline{92.9}  & 93.1  & \textbf{92.5}  & \underline{93.3}  & 79.6  & 76.0  \\
        \rowcolor{Gray}\hspace{-5pt}Hard & (20, 40) & (1s, 3s) & \textbf{93.1}  & \textbf{93.5}  & \underline{92.3}  & \textbf{93.4}  & \textbf{80.9}  & \textbf{77.3}  \\
        \hspace{-5pt}Super-hard & (15, 20) & (0.2s, 1s) & 92.7  & 93.2  & 91.3  & 92.4  & \underline{80.8}  & \underline{77.1} \\
        \end{tabular}
    }
	\end{minipage}
    \hfill
    \begin{minipage}{0.41\linewidth}
            \captionof{table}{Attempts of batch-level perception integration. $\ddagger$ means removing the token normalization term when computing loss.}
            \vspace{-5pt}
    \label{tab:abl_batch}
    \centering
    \renewcommand\arraystretch{1.07}
    \scalebox{0.82}{
        \small
        \begin{tabular}{p{55pt}<{\raggedright}p{16pt}<{\centering}p{16pt}<{\centering}p{16pt}<{\centering}p{16pt}<{\centering}p{16pt}<{\centering}p{20pt}<{\centering}}
        \multirow{2}{*}{\hspace{-5pt}\textbf{Method}} & \multicolumn{2}{c}{\textbf{ID}} & \multicolumn{2}{c}{\textbf{OOD}} & \multicolumn{2}{c}{\textbf{MintVid}} \\
        \cmidrule(lr){2-3} \cmidrule(lr){4-5} \cmidrule(lr){6-7}
        & Acc & F1 & Acc & F1 & Acc & F1 \\
        \shline
        \hspace{-5pt}\gc{Baseline} & \underline{92.9}  & \underline{93.4}  & 91.2  & \underline{92.1}  & 77.8  & 73.3 \\
        \hspace{-5pt}Batch-level & 84.6  & 86.4  & 88.4  & 88.8  & 76.8  & 74.1 \\
        \hspace{-5pt}Batch-level$^\ddagger$ & 89.7  & 90.7  & \underline{91.4}  & \underline{92.1}  & \underline{79.2}  & \underline{76.4} \\
        \rowcolor{Gray}\hspace{-5pt}Phase-level & \textbf{93.1}  & \textbf{93.5}  & \textbf{92.3}  & \textbf{93.4}  & \textbf{80.9}  & \textbf{77.3}  \\
        \rowcolor{Gray}\hspace{-5pt}\textbf{$\bm{\Delta}$ Batch-level$^\ddagger$} & \textcolor{red}{\textbf{+3.4}} & \textcolor{red}{\textbf{+2.8}} & \textcolor{red}{\textbf{+0.9}} & \textcolor{red}{\textbf{+1.3}} & \textcolor{red}{\textbf{+1.7}} & \textcolor{red}{\textbf{+0.9}} \\
        \end{tabular}
    }
	\end{minipage}
    \vspace{-0.2cm}
\end{figure*}

\begin{figure*}[t]
    \centering
    \includegraphics[width=0.98\linewidth]{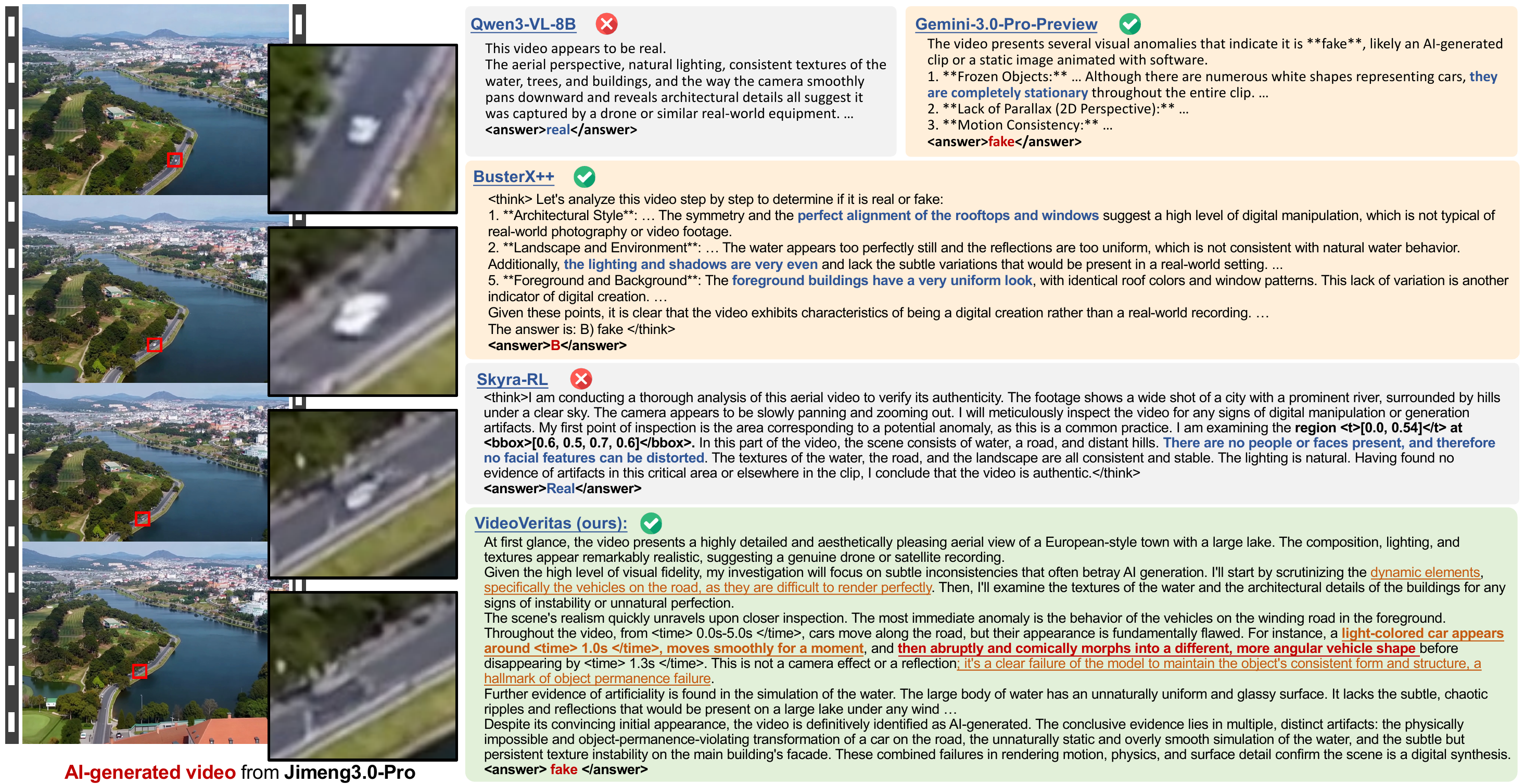}
    \vspace{-0.1cm}
     \caption{\textbf{Comparisons between \videover$\,$ and existing MLLMs.} A small car on the road distorts and disappears.}
	\label{fig:mllm_comp}
    \vspace{-0.2cm}
\end{figure*}

\noindent \textbf{Ablations on PPRL: the quantity of perception data.}
As shown in Figure~\ref{fig:data_ratio}, even introducing a small amount of perception data yields a measurable performance gain (i.e., +1.1\% Acc and 1.3\% F1 on MintVid with $1$K perception data).
As the amount increases, performance improves steadily.
Excessive perception data does not further enhance detection accuracy, but it remains consistently better than the baseline.
A scale around $5$K is efficient and effective to deliver the best performance improvement.

\subsection{Further Analyses}
\label{sec:further_ana}

\noindent \textbf{Difficulty of perception task.}
As mentioned in Sec.~\ref{sec:rl}, the difficulty of SSL task is controllable.
We investigate different settings by adjusting the size and duration of the target objects, where smaller size and shorter exposure times make the task more challenging.
As shown in Table~\ref{tab:abl_task}, increasing the difficulty is beneficial to the detection performance, with the ``Hard'' setting achieving the best results.
Actually, the ``Hard'' setting is already \textit{non-trivial} for humans to perceive, yet the model can still be trained to count them accurately.
This suggests that the perceptual capacity of \textit{machines} may surpass that of \textit{humans}, but remains underexploited.

\noindent \textbf{Why could perception pretext task benefit detection.}
We provide more details about Figure~\ref{fig:reason_ana}.
To facilitate \textit{objective} evaluation, we (1) define five different dimensions and (2) perform pairwise evaluation rather than absolute scoring, where we randomly select $2$K samples ($1.5$K fake and $0.5$K real) that both the baseline and PPRL trained model yield correct answers.
Gemini-3.0-Pro-Preview and GPT-5.1 are adopted for judgment and the results can be ``win'', ``lose'' or ''tie''.
The results show that PPRL trained model tends to analyze \textit{fine-grained entities}, pay more attention to the \textit{object relation} and performs better at \textit{tracking changes}, which consequently benefits the detection.
Evaluation prompts and the definition of the dimensions are listed in Appendix~\ref{sec:app_detail}.

\noindent \textbf{Batch-level integration vs. Phase-level learning.}
Besides phase-level learning,
we also explored batch integration,
where perception and detection task are paired within each mini-batch and trained jointly.
As shown in Table~\ref{tab:abl_batch}, it is non-trivial to directly perform batch integration.
The outputs of detection task are significantly longer than perception counterparts, which limits effective gradients for detection under standard GSPO setting.
We modified the loss computation to balance gradient allocation, which results in certain gains over baseline.
However, phase-level learning remains more effective and is also simpler to apply in practice.

\noindent \textbf{Case studies.}
As compared in Figure~\ref{fig:mllm_comp}, the base model fails to capture fine-grained visual details.
BusterX++ delivers correct answer, but the reasoning is superficial, focusing on elements like the overall context.
Skyra-RL conducts template-like analysis and exhibits suboptimal perception on the high-quality video.
In contrast, \videover$\,$ correctly identifies the distortion of the tiny car, which demonstrates superior fine-grained perception capacities.

\section{Conclusion}
In this paper, we introduced \videover, a framework that integrates fine-grained perception and fact-based reasoning for AI-generated video detection.
We introduce PPRL, which improves perceptual capacities by incorporating simple perception pretext tasks.
We also introduce MintVid, a challenging dataset that contains three evaluation aspects.
Extensive experiments demonstrate the effectiveness of \videover, highlighting the value of learning foundational perceptual skills for complex detection tasks.

\section*{Impact Statement}
This paper presents work whose goal is to advance the field of Machine
Learning. There are many potential societal consequences of our work, none
which we feel must be specifically highlighted here.

\bibliography{main}
\bibliographystyle{icml2026}

%%%%%%%%%%%%%%%%%%%%%%%%%%%%%%%%%%%%%%%%%%%%%%%%%%%%%%%%%%%%%%%%%%%%%%%%%%%%%%%
%%%%%%%%%%%%%%%%%%%%%%%%%%%%%%%%%%%%%%%%%%%%%%%%%%%%%%%%%%%%%%%%%%%%%%%%%%%%%%%
% APPENDIX
%%%%%%%%%%%%%%%%%%%%%%%%%%%%%%%%%%%%%%%%%%%%%%%%%%%%%%%%%%%%%%%%%%%%%%%%%%%%%%%
%%%%%%%%%%%%%%%%%%%%%%%%%%%%%%%%%%%%%%%%%%%%%%%%%%%%%%%%%%%%%%%%%%%%%%%%%%%%%%%
\newpage
\appendix
\onecolumn
\section{More Details of MintVid Dataset}
\label{sec:app_dataset}
\noindent \textbf{Generation pipeline of facial videos.}
To ensure the quality of the generated videos, we carefully curate the input prompts and conditional images.
For the conditional images, we manually select frames from VFHQ~\cite{xie2022vfhq} and HDTF~\cite{zhang2021flow}, ensuring the clearity of selected frames.
For input prompts, we follow the commonly adopted structure, i.e., ``\textbf{detailed caption of the first frame + action control + background description}''.
Specifically, we use Qwen2.5-VL-32B~\cite{bai2025qwen2} to generate detailed caption for the input image.
Then we curate about $20$ action prompts, including control over facial expressions and gestures, e.g., ``slowly turns head from left to right, then briefly dips chin downwards before returning to a neutral position''.
We further curate about $20$ background descriptions, covering different camera controls and environment settings, e.g., ``The camera is fixed, creating a stable frame that emphasizes the subject's presence. The lighting is soft and the background is tastefully blurred.''.
Finally, we utilize Qwen2.5-VL-32B~\cite{bai2025qwen2} to check the plausibility of the prompts, e.g., whether the subject conflicts with the person’s gender in the image and whether background control contradicts the image content.
An example of input prompts is provided in Figure~\ref{prompt:face}.

\begin{figure*}[t]
    \centering
    \includegraphics[width=0.98\linewidth]{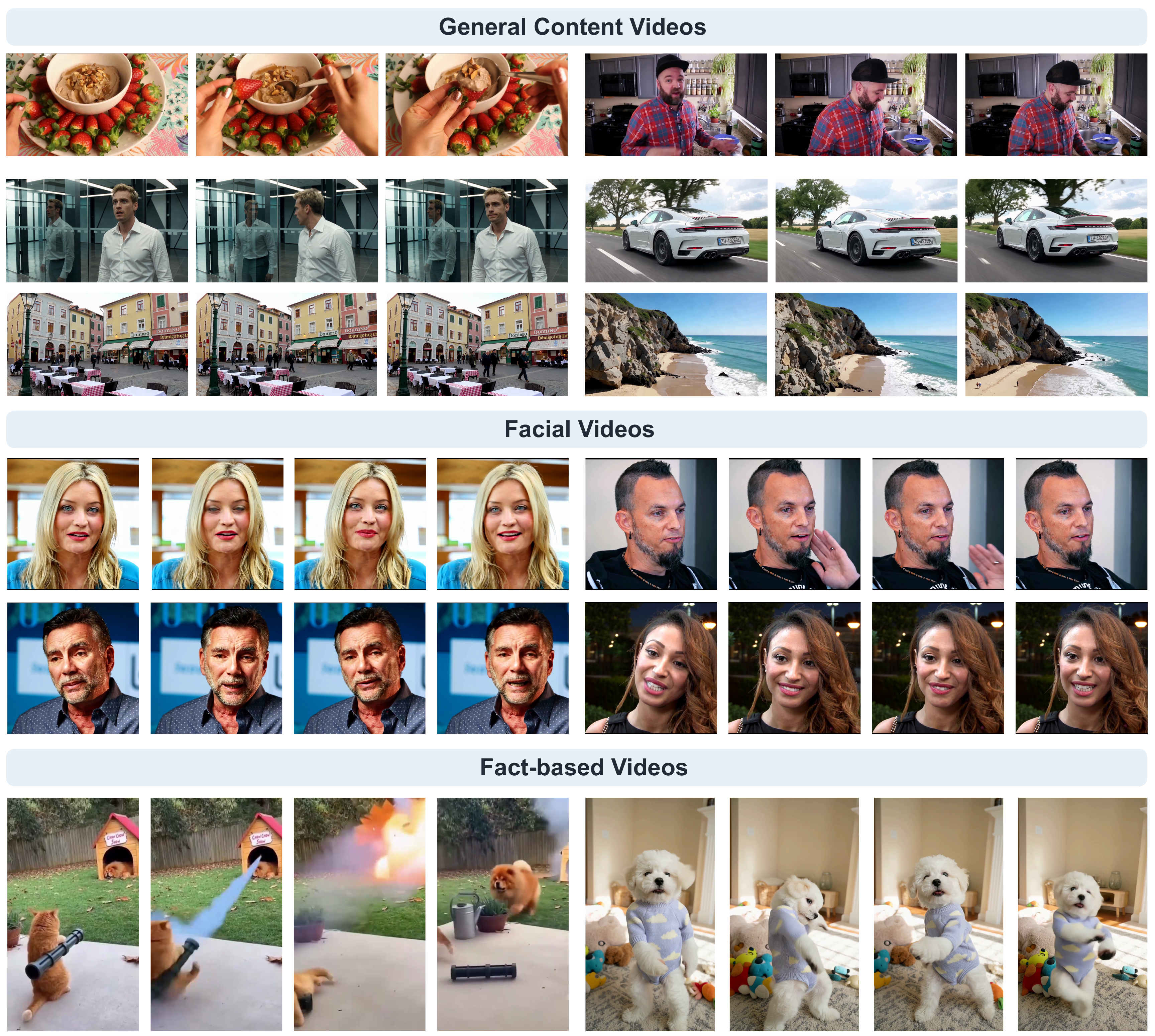}
     \caption{Visual examples of MintVid dataset.}
	\label{fig:supp_mintvid}
    \vspace{-0.4cm}
\end{figure*}

\begin{figure*}[ht]
\centering
\begin{tcolorbox}[listing only, 
                  listing options={basicstyle=\ttfamily\footnotesize},
                  colback=gray!10, 
                  title=An example of input prompt for facial video generation]
A man is shown in a professional setting, likely within a governmental or official building. He is wearing a dark suit, a striped dress shirt, and a patterned tie. His hair is light-colored and neatly combed, and he is wearing rectangular-framed eyeglasses. The background features ornate columns and an American flag, indicating a significant or ceremonial context. He appears to be speaking or giving a statement, though his expression is calm and focused. He brings his left hand to his forehead as if in thought. The lighting mimics the soft, warm glow of the golden hour, casting a flattering light on the man. The ornate columns and American flag remain distinct in the background.
\end{tcolorbox}
\caption{An example of input prompt for facial video generation.}
\vspace{-0.2cm}
\label{prompt:face}
\end{figure*}

\noindent \textbf{Visual examples of MintVid dataset.}
The key attribute of MintVid dataset is that it considers three distinct aspects, aiming to facilitate more complete evaluation.
As mentioned in Sec.~\ref{sec:dataset}, general content videos are generated using $6$ proprietary models.
Facial videos are generated using $3$ specialized public models.
Fact-based videos are collected from online platforms and manually filtered.
We provide more visual examples in Figure~\ref{fig:supp_mintvid}.

\section{Artifacts Taxonomy}
\label{sec:artifact}
To enable more fine-grained analyses, we summarize a taxonomy of common artifacts observed in AI-generated videos,
covering $3$ main perspectives and $11$ detailed aspects.

\textbf{Motion-level Anomalies:}
\begin{itemize}
    \item \textbf{Unnatural Kinematics and Trajectories:} The movement of an object or being defies physical dynamics or biomechanics in its speed, acceleration, or path.
    \item \textbf{Object Permanence Failure:} An object's existence or individuality is violated, causing it to suddenly appear from nowhere, abruptly vanish without reason, illicitly merge with other distinct entities (e.g., two people blending into a single mass), or spontaneously split into multiple copies.
    \item \textbf{Structural Integrity Failure:} A single object, while maintaining its presence, fails to preserve its own physical form, causing its structure or identifying details to illogically warp, stretch, tear, or distort during motion. The object behaves as if made of the wrong material (e.g., a rigid pole bending like rubber).
    \item \textbf{Interaction Anomalies:} Objects violate their physical properties upon contact, or the scene's spatial coherence breaks down during camera movement.
    \item \textbf{Biological Motion Irregularity:} The natural, subtle dynamic features of living beings, such as blinking, breathing, or micro-expressions, are absent, stiff, or unnaturally timed.
\end{itemize}

\textbf{Physical-level Anomalies:}
\begin{itemize}
    \item \textbf{Inconsistent Lighting and Optics:} Shadows, highlights, or reflections behave inconsistently with the scene's established light sources, object geometry, or movement.
    \item \textbf{Causality and Property Violation:} The sequence of events defies cause-and-effect logic, or an object's behavior contradicts its inherent physical properties.
    \item \textbf{Flawed Material Simulation:} The dynamic simulation of complex, non-rigid materials like cloth, smoke, fire, or water appears unrealistic and lacks natural behavior.
    \item \textbf{Contextual and Semantic Mismatch:} The combination of scenes, objects, and actions is logically or commonsensically contradictory, even if individual elements are rendered realistically.
\end{itemize}

\textbf{Perceptual-level Anomalies:}
\begin{itemize}
    \item \textbf{Texture and Surface Instability:} An object's surface texture exhibits a high-frequency flicker, ``crawling'' pattern, or shimmer that is disconnected from its motion.
    \item \textbf{Definition and Clarity Fluctuation:} The sharpness and detail level of a region abruptly and illogically degrade or change without a corresponding cause.
\end{itemize}

\section{Extended Details of Methodology: Prompts and Implementation Details}
\label{sec:app_detail}
\noindent \textbf{Details of reasoning behavior analysis (Figure~\ref{fig:reason_ana}).}
We adopt $5$ objective dimensions to investigate the model's reasoning behavior:
\begin{itemize}
    \item \textbf{Component Granularity:} This dimension evaluates whether the assistant describes the scene as a whole or deconstructs it into specific objects and their fine-grained components (e.g., specific limbs, lens edges, or individual textures).
    \item \textbf{Spatiotemporal Continuity:} This dimension evaluates the assistant's ability to anchor its observations to precise temporal markers (timestamps) and spatial locations. It looks for how well the analysis ``tracks'' changes over a specific timeline.
    \item \textbf{Physics Depth:} This dimension evaluates the extent to which the assistant uses principles of physics (optics, mechanics, biology) to explain anomalies, rather than simply stating that something ``looks wrong''.
    \item \textbf{Forensic Objectivity:} This dimension evaluates the shift from subjective, impression-based language (e.g., ``uncanny'', ``beautiful'') to objective, evidence-based descriptions (e.g., ``static texture overlay'', ``non-uniform deformation'').
    \item \textbf{Relational Logic:} This dimension assesses the analysis of how different elements in the report interact with one another (e.g., the relationship between a moving object and its shadow, or the reaction of a surface to a force).
\end{itemize}
The evaluated samples are correctly answered by both models, which avoids the answer bias.
For each dimension, we compare the outputs of baseline (trained without perception data) and our model (trained with PPRL).
To mitigate potential bias, we utilize two powerful model, i.e., Gemini-3.0-Pro-Preview and GPT-5.1 to conduct pairwise comparison.
The input prompts are presented in Figure~\ref{prompt:reason_ana}.
Specifically, the judging models are required to provide a brief analysis and judgment: model A is better, model B is better or the two models are tied.
Then, we adopt the win rate as a straightforward and unbiased evaluation of the performance.
An example of the judgment is shown below.
\begin{figure*}[ht]
\centering
\begin{tcolorbox}[listing only, 
                  listing options={basicstyle=\ttfamily\footnotesize},
                  colback=gray!10, 
                  title=Judgment Output of ``Component Granularity'']
\{\\
   ``analysis'': ``Assistant A demonstrates superior granular analysis by deconstructing the scene into specific components with detailed observations. \textbf{It isolates and analyzes specific elements} such as \textbf{``small bubbles''} (noting their spherical shape and static nature), distinct light reflections (described as solid, glowing shapes), and the specific ``sun'' object (a glaring red dot) versus the color of the light on the water. \textbf{In contrast, Assistant B focuses on broader features} like the general ``oil slick'', ``edges'', and ``overall scene'', lacking the fine-grained decomposition found in A.'',\\
   ``judgment'': ``[[A]]''\\
\}
\end{tcolorbox}
\caption{Example of jugement output to ``Component Granularity''. The judging model is Gemini-3.0-Pro-Preview.}
\vspace{-0.2cm}
\label{prompt:judge_out}
\end{figure*}

\noindent \textbf{Prompts of \videover.}
Our framework involves perception tasks and detection task, and we do not require careful curation of input prompts.
For detection, the system prompt and input prompt are presented in Figure~\ref{prompt:detection_sys} and Figure~\ref{prompt:detection_user}.
For perception, the system prompt is shared as presented in Figure~\ref{prompt:perception_sys}.
The input instructions are different among different sub-tasks, as shown in Figure~\ref{prompt:grounding_user}, Figure~\ref{prompt:tracking_user}, Figure~\ref{prompt:counting_user} and Figure~\ref{prompt:artifact_user}.

\noindent \textbf{More implementation details.}
Our model is trained with $8$ PPUE GPUs based on ms-swift~\citep{zhao2024swiftascalablelightweightinfrastructure}.
The theoretical peak computational capacity (TFLOPS) of one PPUE GPU is roughly half of an NVIDIA A100 GPU, and each PPUE GPU has $96$GB VRAM.
Suppose the training dataset of RL stage is denoted as $\mathcal{D}$,
the objective function of GSPO is formulated as:
\begin{equation}
\label{eq:gspo}
    \mathcal{L}_{\text{GSPO}}\!=\! -\mathbb{E}_{(\bm{v},\bm{y})\thicksim\mathcal{D}, \{\bm{o}_i\}_{i=1}^G\thicksim\pi_{\theta_{\text{old}}}(\cdot|\bm{v})}\frac{1}{G}\!\sum_{i=1}^G\!\frac{1}{|\bm{o}_i|}\!\sum_{t=1}^{|\bm{o}_i|}\left[\min\left(r_{i}(\theta)A_{i,t}, \text{clip}(r_{i}(\theta),1\!-\!\epsilon_1,1\!+\!\epsilon_2)A_{i,t}\right)\right],
\end{equation}
where importance ratio $r_{i}$ is defined based on sequence likelihood:
\begin{equation}
    r_{i}(\theta)=\left(\!\frac{\pi_\theta(\bm{o}_{i}|\bm{v})}{\pi_{\theta_{\text{old}}}(\bm{o}_{i}|\bm{v})}\!\right)^{\frac{1}{|\bm{o}_i|}},
\end{equation}
and the advantage is calculated based on group estimation:
\begin{equation}
    A_{i,t}=\frac{R_i-\mathrm{mean}(\{R_1,\ldots,R_G\})}{\mathrm{std}(\{R_1,\ldots,R_G\})}.
\end{equation}
Specifically, $G$ is set to $4$ in our method.
$\epsilon_1$ and $\epsilon_2$ are set to $3\times 10^{-4}$ and $4\times 10^{-4}$, respectively.
In the ablations of Table~\ref{tab:abl_batch}, we found that it is non-trivial to perform batch-level perception integration with the above formulation, since the objective is normalized by the output length in Eq.~\ref{eq:gspo}.
To balance the gradient between different tasks, we remove the normalization term $\frac{1}{|\bm{o}_i|}$ in Eq.~\ref{eq:gspo}.
This greatly elevates the performance, which is superior than the baseline, demonstrating that incorporating perception learning is beneficial to our detection task.

\section{More Experimental Results}
\label{sec:more_res}
\noindent \textbf{More metrics.}
As presented in Table~\ref{tab:main_metrics}, we further report recall and F1 (with ``fake'' taken as positive category).
Specifically, NSG-VD exhibits high recall on several subsets, while the overall accuracy is near-chance level.
RestraV suffers from low recall on most datasets.
Zero-shot MLLMs tend to predict videos as real, exhibiting low recall on most datasets.
However, Gemini-2.5-Pro and Gemini-3-Pro-Preview show \textit{clear and decisive leads} in recall, indicating that their perceptual capabilities remain significantly stronger than those of open-source models.
Notably, most models achieve near-perfect results on Emu3, which is collected from MVAD~\cite{hu2025mvad}.
The reason may come from two aspects: (1) most videos in Emu3 are in an anime style, which are generally easy for MLLMs to judge, and (2) the distribution produced by autoregressive generators like Emu3 may be quite different.
BusterX++ performs well on most datasets, but exhibits obvious shortcomings on datasets like OmniAvatar and FantasyTalking.
Skyra-RL fails on most datasets, achieving only 6.9\% averaged recall, which as we mentioned in the main text, may be due to its controlled training procedure.
Overall, \videover$\,$ yields more balanced performance, achieving superior recall and F1 score over previous methods.

\begin{figure*}[t]
    \centering
    \includegraphics[width=0.98\linewidth]{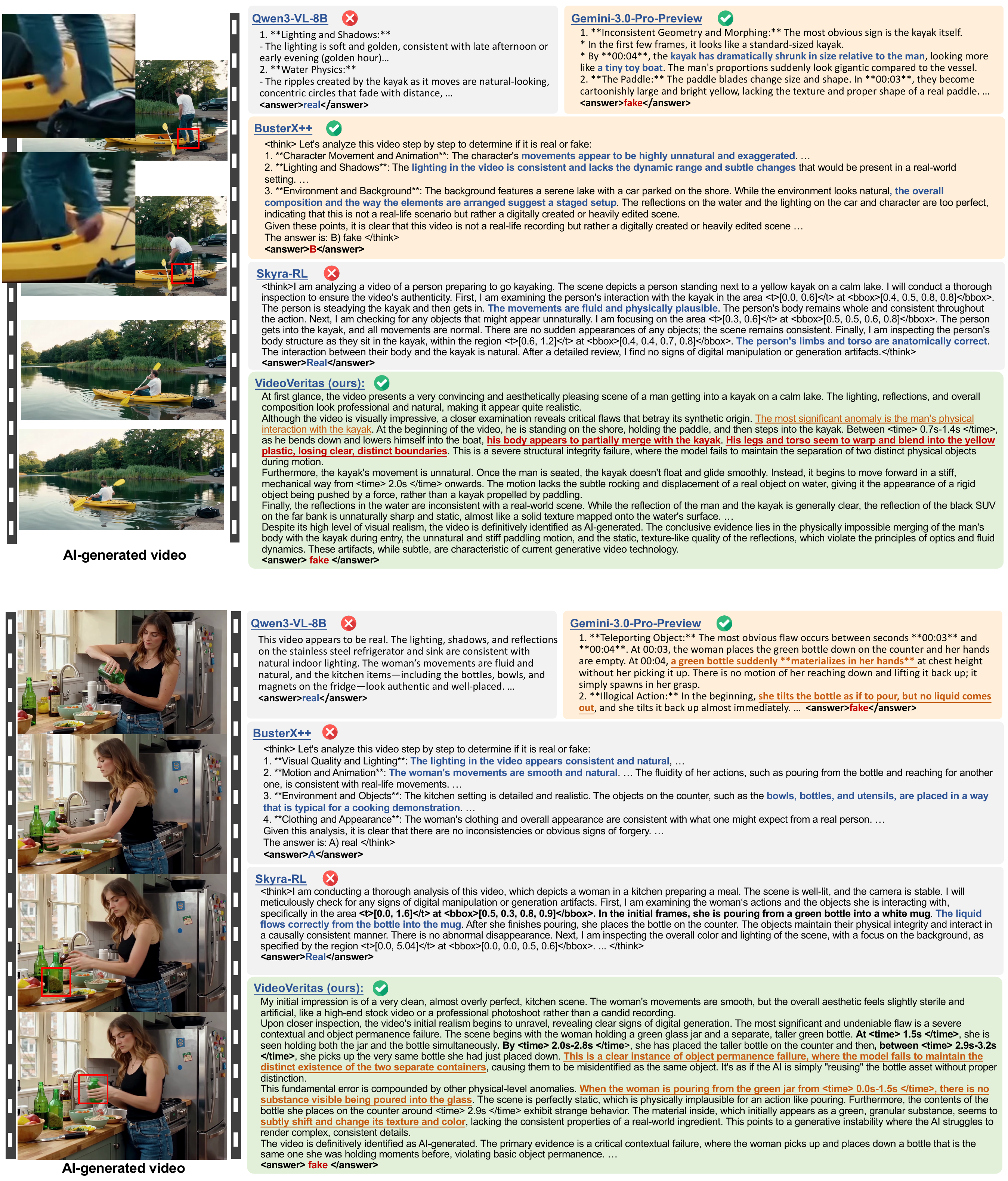}
     \caption{Reasoning comparisons between \videover$\,$ and existing MLLM-based detectors.}
	\label{fig:comp1}
\end{figure*}

\begin{figure*}[t]
    \centering
    \includegraphics[width=0.98\linewidth]{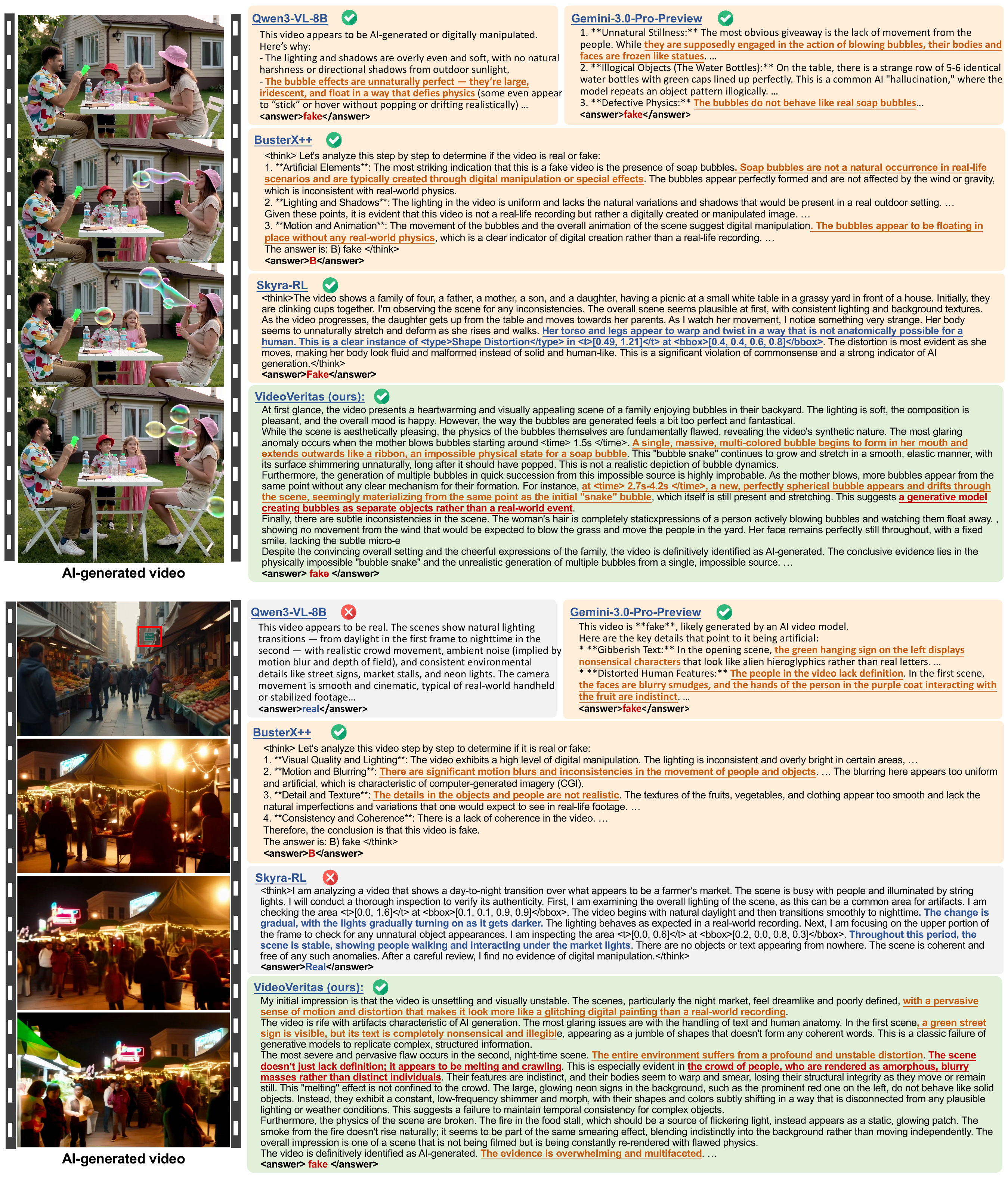}
     \caption{Reasoning comparisons between \videover$\,$ and existing MLLM-based detectors.}
	\label{fig:comp2}
\end{figure*}

\begin{figure*}[t]
    \centering
    \includegraphics[width=0.98\linewidth]{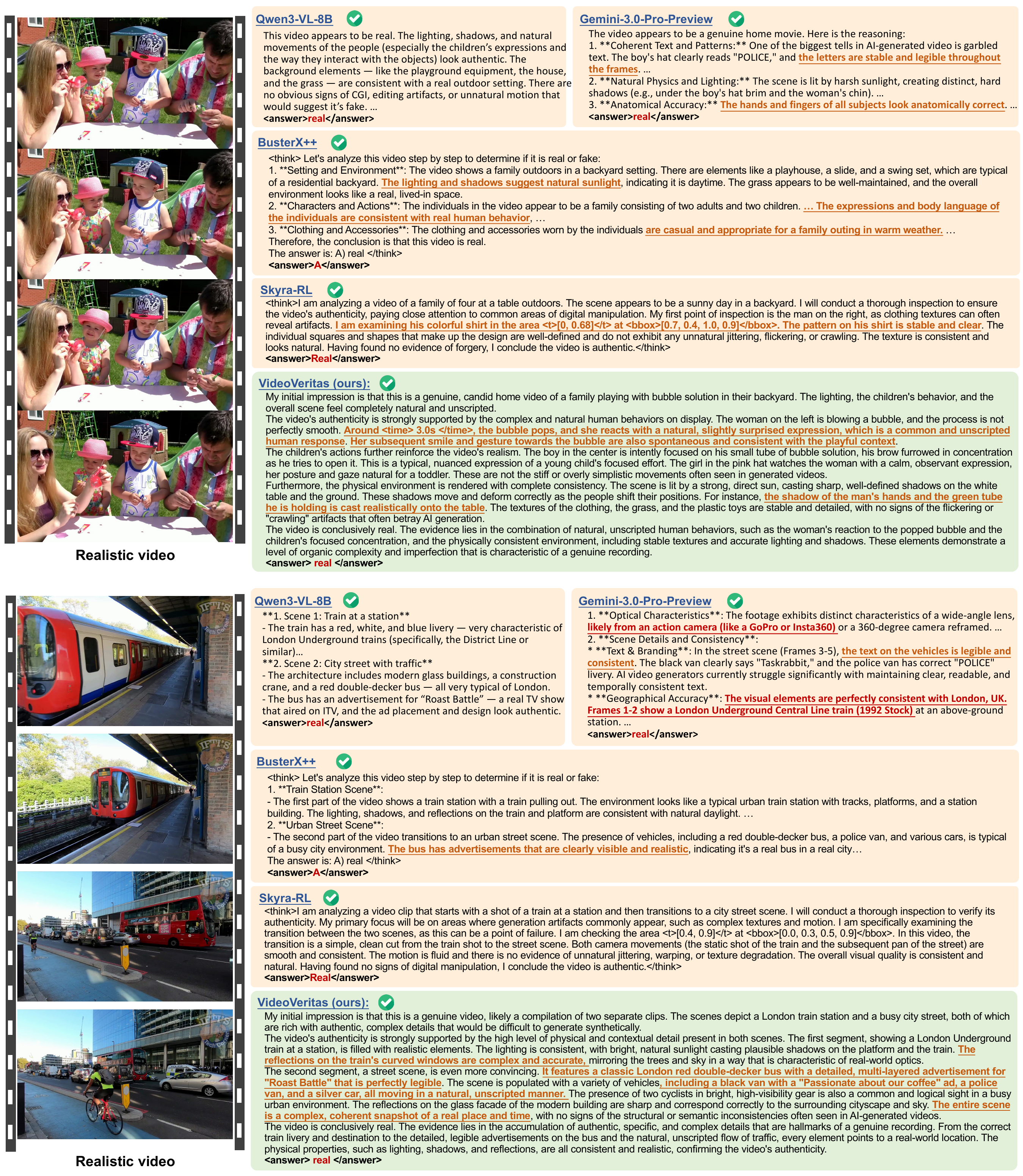}
     \caption{Reasoning comparisons between \videover$\,$ and existing MLLM-based detectors.}
	\label{fig:comp3}
\end{figure*}

\begin{figure*}[t]
    \centering
    \includegraphics[width=0.98\linewidth]{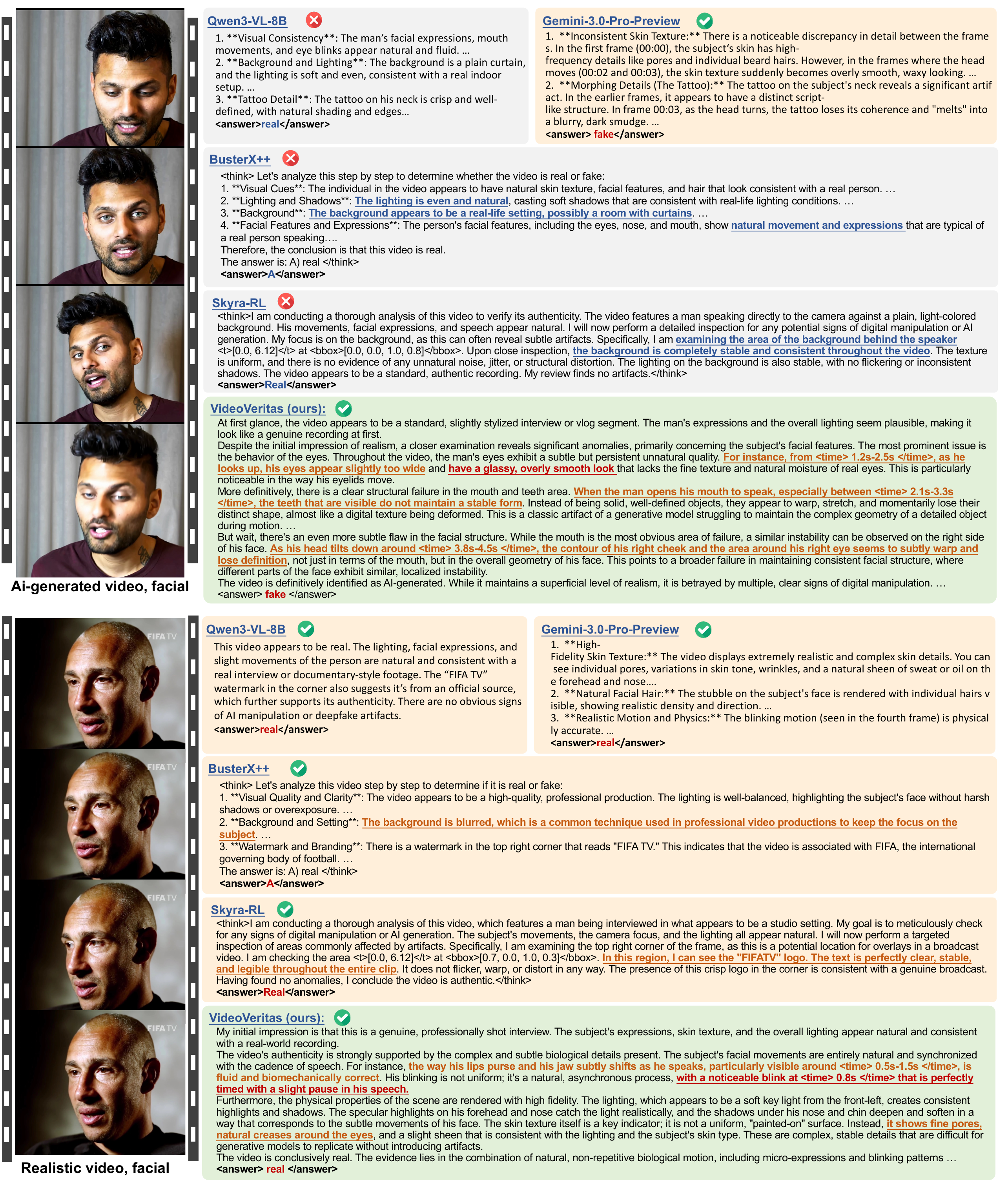}
     \caption{Reasoning comparisons between \videover$\,$ and existing MLLM-based detectors on facial videos.}
	\label{fig:comp4}
\end{figure*}

\begin{figure*}[t]
    \centering
    \includegraphics[width=0.88\linewidth]{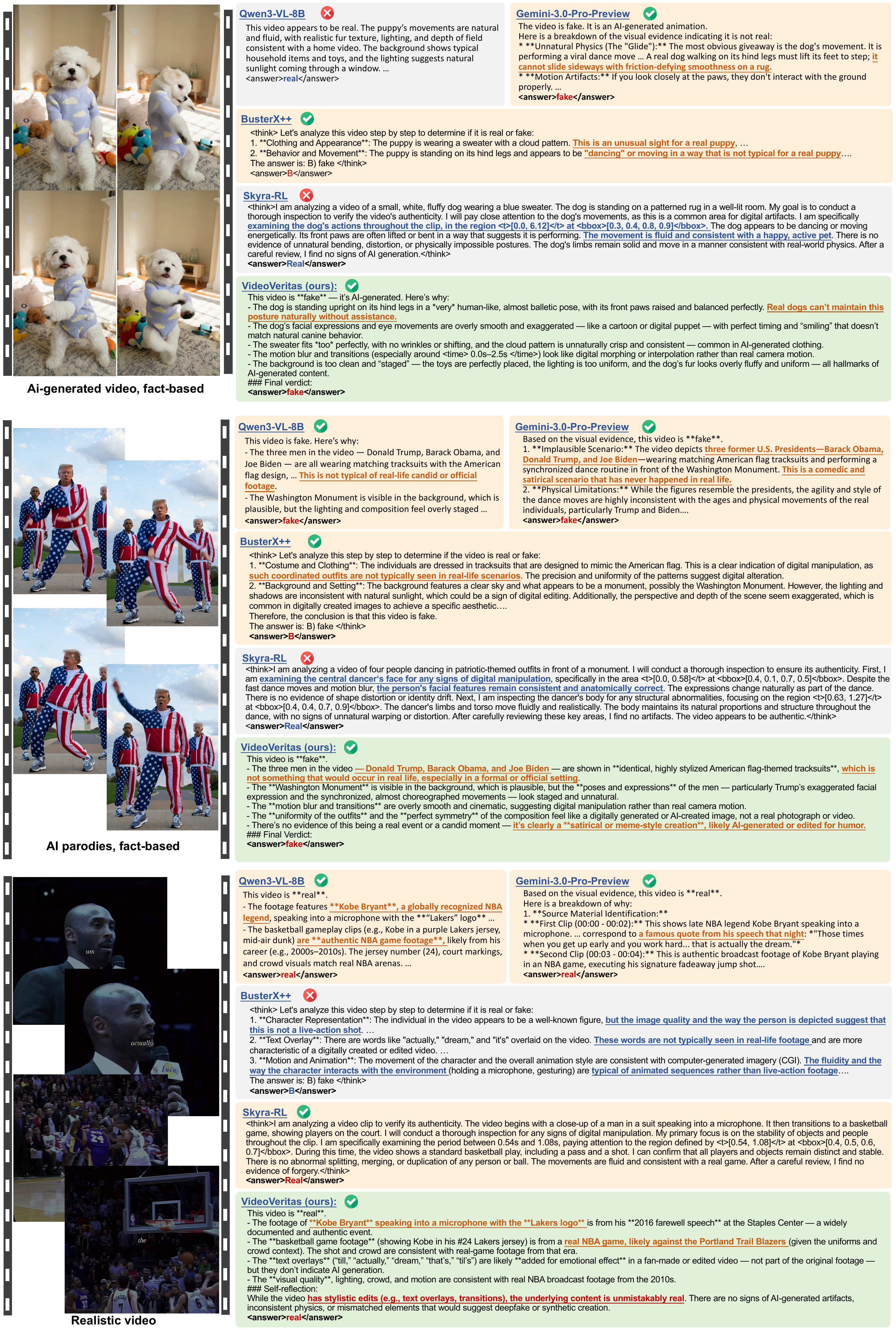}
    \vspace{-0.1cm}
     \caption{Reasoning comparisons between \videover$\,$ and existing MLLM-based detectors on fact-based videos.}
	\label{fig:comp5}
\end{figure*}

\noindent \textbf{More comparisons.}
We provide more comparisons of reasoning outputs in Figure~\ref{fig:comp1}, Figure~\ref{fig:comp2}, Figure~\ref{fig:comp3}, Figure~\ref{fig:comp4} and Figure~\ref{fig:comp5}.
Although Skyra-RL can conduct grounded analysis by examining specific regions, their analyses are mechanical and imprecise, which leads to failures on those fact-based videos, e.g., AI parodies in Figure~\ref{fig:comp5}.
BusterX++ gets correct answers on most samples, but the \textit{reasoning content is superficial}, mainly focus on the macro-level semantic concepts like clothing and lighting.
In contrast, \videover$\,$ demonstrates superior fine-grained perceptual capacity and fact-based reasoning ability.
For instance, in Figure~\ref{fig:comp1} upper, previous methods all fail to perceive the distortion of the person's feet when entering the boat, while \videover$\,$ precisely captures this artifact.
In Figure~\ref{fig:comp1} lower, \videover$\,$ correctly captures the object permanence failure, where a bottle emerges suddenly and the materials in the bottle have unexpectedly changed.

\noindent \textbf{Brief summary of explored attempts.}
We briefly summarize several unsuccessful attempts.
(1) For batch-level integration, we explored conditional reward designs for tight coupling, but it did not achieve improvements comparable to phase-level learning.
(2) We tried more forms of SSL like temporal anomaly grounding (e.g., generated by shuffling local clip), but they were less effective than spatiotemporal tasks like object counting.
(3) We also tried more detailed reward designs for detection task, e.g., explicitly encouraging grounding formats like \texttt{<time>} \texttt{</time>}, which were found to be vulnerable to reward hacking.
However, we suppose the results drawn from such heuristic designs may \textit{differ} when high-quality annotations are available.

\section{Limitation and Future Work}
One limitation of our work is that the perception pretext data used in PPRL mainly comes from broadly available public resources (e.g., generic grounding/tracking and self-supervised counting).
While this choice keeps the training pipeline lightweight and scalable, we do not systematically investigate whether higher-quality, artifact-specific grounding annotations could further improve detection performance and explanation fidelity.
Exploring and curating such high-quality artifact grounding data, as well as integrating it into the pretext RL stage, is a promising direction for future work.

\begin{table*}[t]
\caption{Performance comparison on video datasets, including Accuracy (Acc), Recall (R) and F1. 
$*$ means Average Precision (AP) is adopted following the official guideline. 
The final average performance is calculated by first averaging the results across ID (in-domain), OOD (out-of-domain) and OOD-MintVid, and then taking the average of the three values.
The best results are \textbf{bolded}.}
    \centering
    \renewcommand\arraystretch{1.02}
    \scalebox{0.79}{
        \small
        \begin{tabular}{p{86pt}<{\raggedright}p{15pt}<{\centering}p{15pt}<{\centering}p{15pt}<{\centering}p{15pt}<{\centering}p{15pt}<{\centering}p{15pt}<{\centering}p{15pt}<{\centering}p{15pt}<{\centering}p{15pt}<{\centering}p{15pt}<{\centering}p{15pt}<{\centering}p{15pt}<{\centering}p{15pt}<{\centering}p{15pt}<{\centering}p{15pt}<{\centering}p{15pt}<{\centering}p{15pt}<{\centering}p{20pt}<{\centering}p{16pt}<{\centering}}
        \toprule[1pt]
        \multirow{2}{*}{\hspace{-5pt}\vspace{-2.0em}\textbf{Method}} & \multirow{2}{*}{\vspace{-2.0em}\makebox[0pt][l]{\hspace{-1.5em}\textbf{Metric}}} & \multirow{2}{*}{\vspace{-2.0em}\textbf{ID}} & \multicolumn{6}{c}{\textbf{OOD}} & \multicolumn{10}{c}{\textbf{OOD-MintVid}} & \multirow{2}{*}{\vspace{-2.0em}\textbf{Avg.}} \\
        \cmidrule(lr){4-9} \cmidrule(lr){10-19}
        & & & \rotatebox{40}{GenBuster++} & \rotatebox{40}{LOKI} & \rotatebox{40}{Vidu Q1} & \rotatebox{40}{Gen-4} & \rotatebox{40}{Veo3} & \rotatebox{40}{Emu3} & \rotatebox{40}{Phantom-14B} & \rotatebox{40}{OmniAvatar} & \rotatebox{40}{FantasyTalking}  & \rotatebox{40}{Seedance1.0Pro} & \rotatebox{40}{Jimeng3.0Pro} & \rotatebox{40}{Kling2.5-Turbo} & \rotatebox{40}{Hailuo2.3} & \rotatebox{40}{Wan2.5} & \rotatebox{40}{Sora2} & \rotatebox{40}{Fact} & \\
        \shline
        \rowcolor{lightblue}\textit{\textbf{Small Vision Models}} &&&&&&&&&&&&&&&&&&&\\
        \hspace{-5pt}\rule{0pt}{7pt}D3$^*$ (\textit{ICCV'25}) & AP & 49.5  & 46.8  & 39.1  & 55.1  & 89.0  & 66.1  & 90.0  & 54.8  & 36.4  & 44.1  & 61.6  & 48.7  & 51.7  & 78.9  & 50.5  & 54.6  & 32.4  & 55.1 \\
        \hline
        \multirow{3}{*}{\hspace{-5pt}DeMamba} & Acc & 87.6  & 86.8  & 71.1  & 80.7  & 83.4  & 79.4  & 85.9  & 57.4  & 55.2  & 62.2  & 49.2  & 46.2  & 46.6  & 41.1  & 49.3  & 41.4  & 73.3  & 73.7 \\
        & R & 87.2  & 90.8  & 84.4  & 77.7  & 81.5  & 77.7  & 87.8  & 50.4  & 58.1  & 60.0  & 72.4  & 57.6  & 58.3  & 54.1  & 64.1  & 48.1  & 73.8  & 76.7 \\
        & F1 & 87.3  & 87.3  & 79.5  & 80.1  & 83.1  & 79.1  & 86.1  & 54.2  & 57.9  & 61.3  & 51.1  & 50.6  & 51.2  & 35.1  & 54.8  & 43.0  & 79.5  & 74.6 \\
        \hline
        \multirow{3}{*}{\hspace{-5pt}NSG-VD (\textit{NIPS'25})} & Acc & 52.4  & 53.3  & 55.1  & 53.4  & 58.2  & 64.3  & 50.7  & 54.3  & 53.8  & 58.5  & 53.8  & 58.0  & 56.0  & 69.4  & 57.0  & 57.8  & 52.8  & 55.1 \\
        & R & 28.0  & 76.0  & 75.2  & 60.2  & 84.4  & 87.2  & 1.5  & 48.2  & \textbf{86.4}  & 67.5  & 17.2  & 36.1  & 75.9  & \textbf{83.1}  & 80.9  & \textbf{77.3}  & 30.6  & 50.8 \\
        & F1 & 27.9  & 61.9  & 62.5  & 56.4  & 66.9  & 71.0  & 3.0  & 51.4  & \textbf{65.1}  & 61.9  & 27.1  & 46.3  & 63.3  & 73.2  & 65.3  & \textbf{64.8}  & 39.3  & 45.8 \\
        \hline
        \multirow{3}{*}{\hspace{-5pt}ReStraV (\textit{NIPS'25})} & Acc & 50.7  & 48.9  & 64.8  & 54.8  & 52.8  & 59.8  & 95.2  & 49.6  & 49.3  & 49.9  & 36.9  & 48.4  & 47.2  & 30.2  & 40.9  & 45.4  & 65.0  & 53.2 \\
        & R & 4.3  & 3.8  & 5.6  & 7.8  & 8.0  & 18.2  & 96.2  & 3.7  & 3.7  & 3.7  & 2.5  & 2.5  & 2.5  & 2.5  & 2.5  & 2.5  & 11.1  & 10.4 \\
        & F1 & 8.0  & 6.9  & 9.7  & 14.0  & 14.5  & 30.0  & 95.3  & 6.8  & 6.8  & 6.8  & 4.9  & 4.9  & 4.8  & 4.9  & 4.8  & 4.8  & 16.0  & 14.3 \\
        \shline
        \rowcolor{lightblue}\textit{\textbf{Generic MLLMs}} &&&&&&&&&&&&&&&&&&&\\
        \multirow{3}{*}{\hspace{-5pt}Qwen2.5-VL-7B} & Acc & 54.1  & 59.5  & 49.4  & 58.8  & 53.3  & 60.3  & 99.2  & 50.7  & 50.0  & 49.7  & 63.6  & 49.4  & 51.6  & 53.1  & 55.8  & 50.4  & 87.1  & 57.9 \\
        & R & 13.5  & 23.5  & 27.9  & 15.4  & 8.0  & 18.2  & 99.7  & 1.2  & 0.1  & 0.0  & 14.7  & 6.3  & 7.7  & 8.8  & 19.0  & 6.2  & 84.6  & 20.1 \\
        & F1 & 21.7  & 36.2  & 42.2  & 26.1  & 14.7  & 30.2  & 99.3  & 2.3  & 0.3  & 0.0  & 23.9  & 11.3  & 13.7  & 15.6  & 30.7  & 11.2  & 87.8  & 27.6 \\
        \hline
        \multirow{3}{*}{\hspace{-5pt}Qwen3-VL-8B} & Acc & 65.1  & 62.4  & 56.6  & 60.2  & 59.4  & 62.9  & \textbf{99.9}  & 53.3  & 50.6  & 50.3  & 64.2  & 51.8  & 55.1  & 54.1  & 66.7  & 50.8  & 94.6  & 63.6   \\
        & R & 33.1  & 28.5  & 38.8  & 20.5  & 18.8  & 26.3  & \textbf{100.0}  & 7.0  & 1.4  & 0.9  & 20.7  & 12.5  & 19.2  & 13.0  & 41.6  & 7.4  & 93.6  & 31.2 \\
        & F1 & 44.3  & 43.1  & 54.2  & 34.0  & 31.6  & 41.5  & \textbf{99.9}  & 13.0  & 2.8  & 1.7  & 31.8  & 21.1  & 30.7  & 21.8  & 56.4  & 13.0  & 92.6  & 41.2 \\
        \hline
        \multirow{3}{*}{\hspace{-5pt}InternVL3.5-8B} & Acc & 51.7  & 59.6  & 44.1  & 56.1  & 53.4  & 56.7  & 99.7  & 51.3  & 50.2  & 49.5  & 62.6  & 49.8  & 49.1  & 50.2  & 53.9  & 54.1  & 91.7  & 56.5  \\
        & R & 9.3  & 8.6  & 14.2  & 9.7  & 6.3  & 5.9  & 99.5  & 1.3  & 1.9  & 0.9  & 5.9  & 3.2  & 4.1  & 2.2  & 7.8  & 5.3  & 85.1  & 15.0 \\
        & F1 & 16.3  & 15.5  & 24.5  & 17.3  & 11.9  & 11.2  & 99.7  & 2.5  & 3.8  & 1.8  & 10.9  & 6.2  & 7.8  & 4.3  & 14.2  & 10.0  & 91.4  & 20.5 \\
        \hline
        \multirow{3}{*}{\hspace{-5pt}MiMo-VL-7B} & Acc & 65.8  & 54.3  & 53.6  & 65.5  & 65.8  & 66.0  & 98.8  & 58.4  & 50.1  & 50.6  & 69.7  & 56.7  & 61.7  & 56.1  & 66.0  & 55.4  & 89.9  & 64.8  \\
        & R & 42.4  & 20.3  & 37.0  & 32.6  & 33.2  & 34.0  & \textbf{100.0}  & 21.8  & 5.2  & 6.2  & 33.2  & 21.5  & 31.5  & 16.4  & 39.6  & 16.2  & 95.0  & 38.0 \\
        & F1 & 52.7  & 30.7  & 51.4  & 48.5  & 49.3  & 50.0  & 98.8  & 34.4  & 9.5  & 11.2  & 46.9  & 33.9  & 46.0  & 27.0  & 54.7  & 26.5  & 91.1  & 48.5 \\
        \hline
        \multirow{3}{*}{\hspace{-5pt}Keye-VL-1.5-8B} & Acc & 52.8  & 62.4  & 60.4  & 58.3  & 56.0  & 62.4  & 94.7  & 51.3  & 51.1  & 53.3  & 59.7  & 50.5  & 51.0  & 52.8  & 52.9  & 51.8  & 80.4  & 58.0  \\
        & R & 13.3  & 29.0  & 39.2  & 14.4  & 14.8  & 29.0  & 97.7  & 3.9  & 4.6  & 7.4  & 23.9  & 13.1  & 13.4  & 13.5  & 18.1  & 16.0  & 84.0  & 23.5 \\
        & F1 & 19.9  & 43.8  & 50.4  & 24.7  & 24.6  & 43.8  & 94.0  & 7.3  & 8.7  & 13.6  & 34.7  & 21.2  & 21.5  & 22.2  & 28.0  & 25.2  & 81.3  & 31.1 \\
        \hline
        \multirow{3}{*}{\hspace{-5pt}GLM-4.5V} & Acc & 61.7  & 59.4  & 49.4  & 56.5  & 56.6  & 62.4  & 97.1  & 57.9  & 52.2  & 50.5  & 63.2  & 48.9  & 49.2  & 49.8  & 58.3  & 47.9  & 88.2  & 60.6 \\
        & R & 30.4  & 27.7  & 32.0  & 26.6  & 17.3  & 29.8  & 99.2  & 18.4  & 6.5  & 2.6  & 27.4  & 16.6  & 15.7  & 13.8  & 32.7  & 11.9  & 95.2  & 31.1 \\
        & F1 & 43.3  & 40.6  & 46.4  & 40.4  & 28.5  & 44.2  & 97.1  & 30.3  & 12.0  & 5.0  & 39.2  & 26.8  & 25.5  & 22.9  & 46.7  & 19.9  & 90.1  & 41.6 \\
        \hline
        \multirow{3}{*}{\hspace{-5pt}Qwen3-VL-235B-A22B} & Acc & 66.7  & 64.6  & 57.9  & 57.4  & 65.7  & 63.4  & 98.5  & 54.1  & 50.9  & 51.2  & 65.8  & 49.1  & 32.3  & 49.8  & 63.5  & 50.5  & 93.6  & 63.6  \\
        & R & 38.7  & 36.5  & 44.7  & 29.3  & 31.9  & 28.7  & \textbf{100.0}  & 9.1  & 2.6  & 3.3  & 32.3  & 16.2  & 32.3  & 13.7  & 41.3  & 16.2  & 94.3  & 36.7 \\
        & F1 & 51.6  & 50.8  & 59.4  & 45.0  & 48.1  & 43.8  & 98.5  & 16.5  & 4.9  & 6.3  & 45.1  & 26.4  & 48.8  & 22.7  & 55.8  & 26.2  & 94.3  & 48.0 \\
        \hline
        \multirow{3}{*}{\hspace{-5pt}Gemini-2.5-Pro} & Acc & 80.5  & 71.8  & 70.2  & 70.8  & 76.0  & 70.4  & 93.3  & 57.8  & 55.8  & 55.6  & 69.4  & 60.2  & 65.6  & 62.7  & 65.7  & 60.9  & 88.3  & 73.4 \\
        & R & 85.1  & 59.8  & 65.5  & 38.2  & 65.6  & 46.8  & \textbf{100.0}  & 28.2  & 22.6  & 18.2  & 52.0  & 34.9  & 47.7  & 39.4  & 47.7  & 39.4  & 95.5  & 63.4 \\
        & F1 & 81.4  & 68.1  & 70.3  & 55.3  & 74.7  & 60.4  & 93.7  & 40.6  & 33.8  & 27.9  & 60.9  & 45.9  & 58.5  & 51.0  & 58.4  & 51.0  & 90.3  & 67.9 \\
        \hline
        \multirow{3}{*}{\hspace{-5pt}Gemini-3-Pro-Preview} & Acc & 79.4  & 82.5  & 74.9  & 83.1  & 76.3  & 85.7  & 89.3  & 65.5  & \textbf{65.3}  & 61.5  & 81.4  & 72.1  & 73.1  & 74.3  & 73.1  & 63.5  & 91.4  & 77.8 \\
        & R & 96.2  & 93.5  & \textbf{94.4}  & \textbf{92.7}  & 75.8  & 89.6  & \textbf{100.0}  & 60.8  & 60.2  & 52.7  & \textbf{81.2}  & \textbf{71.8}  & 73.9  & 76.0  & 73.9  & 54.4  & 98.6  & 86.3 \\
        & F1 & 82.7  & 84.3  & 80.2  & 84.7  & 76.1  & 86.3  & 90.3  & 63.9  & 63.4  & 57.6  & 81.3  & 71.9  & 73.1  & 74.5  & 73.3  & 59.7  & 92.6  & 79.1 \\
        \shline
        \rowcolor{lightblue}\multicolumn{4}{c}{\hspace{-22pt}\textit{\textbf{MLLM-based Video Forgery Detectors}}} &&&&&&&&&&&&&&&&\\
        \multirow{3}{*}{\hspace{-5pt}BusterX++} & Acc & 77.1  & 79.6  & 70.9  & 53.0  & 89.9  & 76.2  & 98.9  & 62.3  & 48.6  & 47.6  & 71.8  & 71.7  & 82.3  & 70.4  & \textbf{87.3}  & 58.7  & 88.8  & 74.7  \\
        & R & 64.9  & 68.8  & 67.5  & 6.1  & 82.9  & 55.5  & \textbf{100.0}  & 30.6  & 3.3  & 1.3  & 47.8  & 47.3  & 73.9  & 45.6  & \textbf{86.2}  & 13.0  & 90.0  & 57.4 \\
        & F1 & 71.4  & 77.1  & 75.5  & 11.4  & 89.2  & 70.0  & 98.9  & 44.8  & 6.0  & 2.4  & 57.7  & 57.2  & 77.1  & 55.6  & 84.5  & 19.8  & 89.7  & 63.7 \\
        \hline
        \multirow{3}{*}{\hspace{-5pt}Skyra-RL} & Acc & 52.1  & 55.7  & 37.7  & 52.9  & 59.1  & 55.9  & 51.5  & 50.1  & 49.9  & 49.9  & 63.4  & 51.7  & 53.5  & 51.7  & 59.6  & 50.6  & 51.9  & 52.5  \\
        & R & 4.6  & 11.7  & 7.3  & 6.1  & 18.2  & 12.1  & 3.0  & 0.3  & 0.0  & 0.0  & 9.9  & 6.9  & 10.8  & 2.7  & 22.3  & 1.0  & 11.4  & 6.9 \\
        & F1 & 8.4  & 20.9  & 13.5  & 11.4  & 30.7  & 21.5  & 5.8  & 0.6  & 0.0  & 0.0  & 17.9  & 12.8  & 19.4  & 5.3  & 36.4  & 2.0  & 20.5  & 12.4 \\
        \hline
        \rowcolor{Gray}\multirow{3}{*}{\hspace{-7pt} \videover $\,$(\textbf{ours})} & Acc & \textbf{93.1}  & \textbf{91.4}  & \textbf{78.1}  & \textbf{93.6}  & \textbf{96.7}  & \textbf{94.5}  & 99.4  & \textbf{79.0}  & 56.2  & \textbf{84.5}  & \textbf{86.6}  & \textbf{80.8}  & \textbf{86.0}  & \textbf{85.8}  & 86.3  & \textbf{67.6}  & \textbf{96.1}  & \textbf{88.8} \\
        & R & \textbf{99.3}  & \textbf{94.9}  & 91.0  & 89.5  & \textbf{96.0}  & \textbf{92.3}  & \textbf{100.0}  & \textbf{76.6}  & 30.5  & \textbf{87.5}  & 75.7  & 68.4  & \textbf{78.6}  & 78.1  & 79.3  & 41.1  & \textbf{100.0}  & \textbf{88.3} \\
        & F1 & \textbf{93.5}  & \textbf{91.7}  & \textbf{84.7}  & \textbf{93.3}  & \textbf{96.7}  & \textbf{94.3}  & 99.4  & \textbf{78.5}  & 40.9  & \textbf{84.9}  & \textbf{82.0}  & \textbf{78.6}  & \textbf{85.4}  & \textbf{84.4}  & \textbf{85.8}  & 55.8  & \textbf{96.6}  & \textbf{88.0} \\
        \bottomrule[1pt]
        \end{tabular}
    }
    \label{tab:main_metrics}
\end{table*}

\begin{table*}[t]
\caption{Detailed ablation of perception tasks on MindVid dataset. The reported metrics are ``Acc/Recall/F1''.}
    \centering
    \renewcommand\arraystretch{1.1}
    \scalebox{0.81}{
        \small
        \begin{tabular}{p{70pt}<{\raggedright}p{46pt}<{\centering}p{46pt}<{\centering}p{46pt}<{\centering}p{46pt}<{\centering}p{46pt}<{\centering}p{46pt}<{\centering}p{46pt}<{\centering}p{46pt}<{\centering}p{46pt}<{\centering}}
        \toprule[1pt]
        \hspace{-5pt}\textbf{Task} & \rotatebox{40}{\textbf{Phantom-14B}} & \rotatebox{40}{\textbf{OmniAvatar}} & \rotatebox{40}{\textbf{FantasyTalking}} & \rotatebox{40}{\textbf{Seedance1.0Pro}} & \rotatebox{40}{\textbf{Jimeng3.0Pro}} & \rotatebox{40}{\textbf{Kling2.5-Turbo}} & \rotatebox{40}{\textbf{Hailuo2.3}} & \rotatebox{40}{\textbf{Wan2.5}} & \rotatebox{40}{\textbf{Sora2}} \\
        \shline
        \hspace{-5pt}\rule{0pt}{8pt}w/o Perception & 77.1/76.0/76.8 & 52.6/27.1/36.4 & 84.7/91.2/85.6 & 81.3/67.5/74.5 & 76.2/62.6/73.1 & 83.5/76.9/82.9 & 82.8/75.3/81.2 & 83.0/75.9/82.2 & 61.1/31.3/44.5 \\
        \hspace{-5pt}SSL & \textbf{81.0}/\textbf{86.1}/\textbf{81.9} & \textbf{62.5}/\textbf{49.1}/\textbf{56.7} & 85.5/\textbf{95.2}/86.8 & 84.1/74.3/78.9 & 78.9/67.9/76.9 & 83.7/77.2/83.1 & 82.4/78.1/81.4 & 84.6/78.9/84.1 & 65.5/40.1/53.6 \\
        \hspace{-5pt}G-G & 76.2/72.3/75.3 & 58.2/36.2/46.4 & 84.2/88.2/84.8 & 85.3/76.4/80.7 & 79.4/68.2/77.4 & 85.1/79.3/84.7 & \textbf{86.1}/\textbf{82.9}/\textbf{85.5} & 83.9/77.1/83.3 & 66.5/41.4/55.2 \\
        \hspace{-5pt}A-G & 79.5/79.7/79.5 & 58.2/37.2/47.1 & 85.2/91.2/86.1 & 84.5/73.9/79.4 & 79.4/67.8/77.2 & 84.3/77.5/83.7 & \textbf{86.1}/82.2/85.4 & 84.4/77.6/83.8 & \textbf{68.7}/\textbf{45.5}/\textbf{59.1} \\
        \hspace{-5pt}SSL $+$ G-G & 79.0/76.6/78.5 & 56.0/30.5/40.9   & 84.5/87.5/84.9 & \textbf{86.6}/75.7/82.0 & \textbf{80.8}/68.4/\textbf{78.6} & \textbf{86.0}/78.6/\textbf{85.4} & 85.8/78.1/84.4 & \textbf{86.3}/\textbf{79.3}/\textbf{85.8} & 67.6/41.1/55.8 \\
        \hspace{-5pt}SSL $+$ A-G & 79.8/80.6/79.9 & 58.8/38.6/48.4 & 86.2/93.3/87.1 & 82.7/70.0/76.5 & 78.1/65.7/75.6 & 83.8/76.9/83.1 & 83.8/78.1/82.6 & 83.6/76.5/82.9 & 64.2/36.7/50.5 \\
        \hspace{-5pt}G-G $+$ A-G & 78.1/80.4/78.6 & 57.1/38.4/47.3 & 84.9/94.0/86.2 & 86.4/\textbf{79.6}/\textbf{82.5} & 79.7/\textbf{69.2}/77.9 & 85.1/\textbf{79.6}/84.7 & 85.1/81.5/84.4 & 84.8/78.9/84.3 & 63.7/36.0/49.7 \\
        \hspace{-5pt}SSL $+$ G-G $+$ A-G & 79.1/76.9/78.7 & 60.1/38.9/49.4 & \textbf{87.8}/94.2/\textbf{88.5} & 81.3/68.5/74.7 & 74.4/59.8/70.7 & 81.9/74.4/81.0 & 81.1/75.3/79.7 & 83.3/77.1/82.7 & 64.2/38.0/51.4 \\
        \bottomrule[1pt]
        \end{tabular}
    }
    \label{tab:task_mintvid}
\end{table*}

\clearpage

\begin{figure*}[t]
\centering
\begin{tcolorbox}[listing only, 
                  listing options={basicstyle=\ttfamily\footnotesize},
                  colback=gray!10, 
                  title=System Prompt for Detection]
You are an expert video analyst.\\
Please think about the question as if you were a human pondering deeply. It’s encouraged to include self-reflection or verification in the reasoning process. Then, give a final verdict within \texttt{<answer>} \texttt{</answer>} tags.
\end{tcolorbox}
\caption{System prompt for detection task.}
\label{prompt:detection_sys}
\end{figure*}

\begin{figure*}[t]
\centering
\begin{tcolorbox}[listing only, 
                  listing options={basicstyle=\ttfamily\footnotesize},
                  colback=gray!10, 
                  title=User Prompt for Detection]
\texttt{<video>}\\
Is this video real or fake?
\end{tcolorbox}
\caption{User prompt for detection task.}
\label{prompt:detection_user}
\end{figure*}

\begin{figure*}[t]
\centering
\begin{tcolorbox}[listing only, 
                  listing options={basicstyle=\ttfamily\footnotesize},
                  colback=gray!10, 
                  title=System Prompt for Perception]
You are an expert video analyst. Based on this video, provide the answer directly.
\end{tcolorbox}
\caption{System prompt for perception task.}
\label{prompt:perception_sys}
\end{figure*}

\begin{figure*}[t]
\centering
\begin{tcolorbox}[listing only, 
                  listing options={basicstyle=\ttfamily\footnotesize},
                  colback=gray!10, 
                  title=User Prompt for Spatiotemporal Grounding]
\texttt{<video>}\\
Given the query ``\texttt{[description]}'', when and where does the described content occur in the video? please firstly give the start and end time, spatial bounding box corresponding to each integer second.\\
\\
Please provide only the time span in seconds and bounding boxes as JSON, ONLY up to 16 seconds.\\
You MUST output one bounding box for every integer second within the given time span (inclusive).\\
Example:\\
\{``time'': [8.125, 13.483], ``boxes'': \{``9'': [317, 422, 582, 997], ``10'': [332, 175, 442, 369], ``11'': [340, 180, 450, 370]\}\}\\
Note: Each key in ``boxes'' must be an integer second within the span, and its value must be a 4-number bounding box [x1, y1, x2, y2].
\end{tcolorbox}
\caption{User prompt for spatiotemporal grounding task. ``\texttt{[description]}'' is a brief caption of the target object, which is included in the dataset, such as ``a baby closes a laptop in the study''.}
\label{prompt:grounding_user}
\end{figure*}

\begin{figure*}[t]
\centering
\begin{tcolorbox}[listing only, 
                  listing options={basicstyle=\ttfamily\footnotesize},
                  colback=gray!10, 
                  title=User Prompt for Object Tracking]
\texttt{<video>}\\
Given the bounding box ``\texttt{[initial box]}'' of the target object in the first frame, track this object in each frame and output its bounding box once per second.ONLY up to 16 seconds.\\
Example:\\
\{``boxes'': \{``1'': [405, 230, 654, 463], ``2'': [435, 223, 678, 446], ..., ``16'': [415, 203, 691, 487]\}\}\\
Note: Each key in ``boxes'' must correspond to a second (1, 2, 3, ..., 16) and contain a 4-number bounding box [x1, y1, x2, y2].
\end{tcolorbox}
\caption{User prompt for object tracking task. ``\texttt{[initial box]}'' is the bounding box in the first frame, which is included in the dataset.}
\label{prompt:tracking_user}
\end{figure*}

\begin{figure*}[t]
\centering
\begin{tcolorbox}[listing only, 
                  listing options={basicstyle=\ttfamily\footnotesize},
                  colback=gray!10, 
                  title=User Prompt for Object Counting]
\texttt{<video>}\\
Count the number of circles, squares, and triangles that appear in this video. Be aware that the shapes can appear in any color and at any angle of rotation. They may be present on one or multiple frames, and any given frame can contain more than one shape. Provide the answer as three comma-separated numbers in the format: circles,squares,triangles. For example, if you see 3 circles, 1 square, and 4 triangles, your answer should be ``3,1,4''.
\end{tcolorbox}
\vspace{-0.3cm}
\caption{User prompt for object counting task.}
\vspace{-0.2cm}
\label{prompt:counting_user}
\end{figure*}

\begin{figure*}[t]
\centering
\begin{tcolorbox}[listing only, 
                  listing options={basicstyle=\ttfamily\footnotesize},
                  colback=gray!10, 
                  title=User Prompt for Artifact Grounding]
\texttt{<video>}\\
Find the visual artifacts at ``\texttt{[time]}'' in the video.\\
Provide the bounding boxes where the artifact occurred, in [x$_{min}$, y$_{min}$, x$_{max}$, y$_{max}$] format. If there are multiple locations, you should provide all the bounding boxes.\\
Example:\\
\{``boxes'': [[487, 324, 573, 398], [670, 533, 734, 769], ...]]\}.
\end{tcolorbox}
\vspace{-0.3cm}
\caption{User prompt for artifact grounding task (as mentioned in Sec.~\ref{sec:ablation}). ``\texttt{[time]}'' is the exact target timestamp in seconds, which is included in the dataset, such as ``4.5s''.}
\vspace{-0.2cm}
\label{prompt:artifact_user}
\end{figure*}

\begin{figure*}[t]
\centering
\begin{tcolorbox}[listing only, 
                  listing options={basicstyle=\ttfamily\footnotesize},
                  colback=gray!10, 
                  title=Input Prompt for Reasoning Behavior Evaluation]
You are a helpful assistant proficient in analyzing vision reasoning problems.\\
\#\# Instruction:\\
You will be presented with two analytical reports (Assistant A and Assistant B) that describe observations from a video. Your task is to perform a side-by-side comparison and determine which assistant provides higher-quality reasoning based \textbf{ONLY} on the specific dimension provided below.\\
The evaluation must be conducted strictly based on the textual evidence provided. Do not assume any external video information. Your goal is to identify which assistant demonstrates more professional, precise, and logically structured analysis within the specified scope.\\
\\
\#\# Evaluation Dimension: \texttt{\{Dimension Name\}}\\
\textbf{Description}: \texttt{\{Dimension Description\}}\\
\\
\#\# Rules for Evaluation:\\
- \textbf{Strictly Dimension-Focused}: Ignore other aspects of the reports. Only judge based on the provided dimension.\\
- \textbf{Content over Conclusion}: Do not favor an assistant based on its final verdict. Focus on the depth and quality of the reasoning path.\\
- \textbf{Neutrality}: Be unbiased toward length or tone. Prioritize the density of meaningful, analytical information.\\
\\
\#\# Desired Output Format:\\
Present your verdict in a JSON format, with key ``analysis'' for a short reason of your judgment and key ``judgment'' to indicate your decision: use ``[[A]]'' if assistant A prevails, ``[[B]]'' if assistant B does, and ``[[C]]'' for a tie. \\
\\
\#\# Input Data:\\
\lbrack The Start of Assistant A’s Analysis\rbrack\\
\texttt{\{Output of Assistant A\}}\\
\lbrack The End of Assistant A’s Analysis\rbrack\\
\lbrack The Start of Assistant B’s Analysis\rbrack\\
\texttt{\{Output of Assistant B\}}\\
\lbrack The End of Assistant B’s Analysis\rbrack\\
\end{tcolorbox}
\vspace{-0.2cm}
\caption{Input Prompt for Reasoning Behavior Evaluation (Figure~\ref{fig:reason_ana}).}
\label{prompt:reason_ana}
\end{figure*}

% You can have as much text here as you want. The main body must be at most $8$
% pages long. For the final version, one more page can be added. If you want, you
% can use an appendix like this one.

% The $\mathtt{\backslash onecolumn}$ command above can be kept in place if you
% prefer a one-column appendix, or can be removed if you prefer a two-column
% appendix.  Apart from this possible change, the style (font size, spacing,
% margins, page numbering, etc.) should be kept the same as the main body.
%%%%%%%%%%%%%%%%%%%%%%%%%%%%%%%%%%%%%%%%%%%%%%%%%%%%%%%%%%%%%%%%%%%%%%%%%%%%%%%
%%%%%%%%%%%%%%%%%%%%%%%%%%%%%%%%%%%%%%%%%%%%%%%%%%%%%%%%%%%%%%%%%%%%%%%%%%%%%%%

\end{document}